\documentclass[sigconf]{acmart}
\AtBeginDocument{%
  }

\copyrightyear{2023} 
\acmYear{2023} 
\setcopyright{acmlicensed}\acmConference[KDD '23]{Proceedings of the 29th ACM SIGKDD Conference on Knowledge Discovery and Data Mining}{August 6--10, 2023}{Long Beach, CA, USA}
\acmBooktitle{Proceedings of the 29th ACM SIGKDD Conference on Knowledge Discovery and Data Mining (KDD '23), August 6--10, 2023, Long Beach, CA, USA}
\acmPrice{15.00}
\acmDOI{10.1145/3580305.3599373}
\acmISBN{979-8-4007-0103-0/23/08}

\newcommand{\specialcell}[2][c]{%
	\begin{tabular}[#1]{@{}c@{}}#2\end{tabular}}

\usepackage[ruled]{algorithm2e}
\usepackage{multirow}
\usepackage{arydshln}
\usepackage{ulem}
\usepackage{subfigure}
\usepackage{booktabs}

\usepackage{amsmath}
\usepackage{pifont}

\usepackage{colortbl}
\usepackage{balance}
\usepackage{flushend}

\definecolor{lightgrey}{rgb}{0.9,0.9,0.9}

\DeclareMathOperator*{\argmin}{arg\,min}

\newcommand{\bb}{\hspace{-1mm} $\bullet$}
\newcommand{\mymodel}{GraphGLOW\xspace}
\newcommand{\mymodelvar}{GraphGLOW*\xspace}
\newcommand{\mymodelgat}{$\mathrm{GraphGLOW_{at}}$\xspace}

\begin{document}
	
        \title{GraphGLOW: Open-World Graph Structure Learning for Cross-Graph Topology Refinement}
        \title{GraphGLOW: Universal and Generalizable Structure Learning for Graph Neural Networks}
	
	\author{Wentao Zhao}
	\email{permanent@sjtu.edu.cn}
	\affiliation{%
		\institution{Shanghai Jiao Tong University}
		\city{Shanghai}
		\country{China}
	}
	
	\author{Qitian Wu}
	\email{echo740@sjtu.edu.cn}
	\affiliation{%
		\institution{Shanghai Jiao Tong University}
		\city{Shanghai}
		\country{China}
	}
	
	\author{Chenxiao Yang}
	\email{chr26195@sjtu.edu.cn}
	\affiliation{%
		\institution{Shanghai Jiao Tong University}
		\city{Shanghai}
		\country{China}
	}
	
	\author{Junchi Yan}
        \authornote{Corresponding author.}
	\email{yanjunchi@sjtu.edu.cn}
	\affiliation{%
		\institution{Shanghai Jiao Tong University}
		\city{Shanghai}
		\country{China}
	}
	

	\renewcommand{\shortauthors}{Wentao Zhao, Qitian Wu, Chenxiao Yang, \& Junchi Yan}
	
	\begin{abstract}
	Graph structure learning is a well-established problem that aims at optimizing graph structures adaptive to specific graph datasets to help message passing neural networks (i.e., GNNs) to yield effective and robust node embeddings. However, the common limitation of existing models lies in the underlying \textit{closed-world assumption}: the testing graph is the same as the training graph. This premise requires independently training the structure learning model from scratch for each graph dataset, which leads to prohibitive computation costs and potential risks for serious over-fitting. To mitigate these issues, this paper explores a new direction that moves forward to learn a universal structure learning model that can generalize across graph datasets in an open world. We first introduce the mathematical definition of this novel problem setting, and describe the model formulation from a probabilistic data-generative aspect. Then we devise a general framework that coordinates a single graph-shared structure learner and multiple graph-specific GNNs to capture the generalizable patterns of optimal message-passing topology across datasets. The well-trained structure learner can directly produce adaptive structures for unseen target graphs without any fine-tuning. Across diverse datasets and various challenging cross-graph generalization protocols, our experiments show that even without training on target graphs, the proposed model i) significantly outperforms expressive GNNs trained on input (non-optimized) topology, and ii) surprisingly performs on par with state-of-the-art models that independently optimize adaptive structures for specific target graphs, with notably orders-of-magnitude acceleration for training on the target graph.
	\end{abstract}

	
	
	
	
	\maketitle
	
	\section{Introduction}\label{sec-intro}
	Graph neural networks (GNNs)~\cite{gcn, gat, graphsage}, as a de facto model class based on the message passing principle, have shown promising efficacy for learning node representations for graph-structured data, with extensive applications to, e.g., physics simulation~\cite{sanchez2020learning}, traffic prediction~\cite{jiang2022graph}, drug recommendation~\cite{yang2023molerec}. However, due to the inevitable error-prone data collection~\cite{idgl}, the input graph may contain spurious and unobserved edges that lead to sub-optimal results of GNNs and degrade the downstream performance.
 
 Graph structure learning~\cite{structure_learning_survey} serves as a plausible remedy for such an issue via optimizing graph structures and GNN classifiers at the same time. To this end, recent endeavors explore different technical aspects, e.g., parameterizing each potential edge between any pair of nodes~\cite{lds,variational_inference_2} or estimating potential links through a parameterized network~\cite{idgl,wunodeformer,rbf},  etc. 
However, existing models limit their applicability within a closed-world hypothesis: the training and testing of structure learning models, which optimize the graph structures, are performed on the same graph. The issue, however, is that since structure learning is often heavy-weighted and requires sophisticated optimization, it can be prohibitively resource-consuming to train structure learning models from scratch for each graph dataset. Moreover, due to limited labels in common graph-based predictive tasks, structure learning models are prone to over-fitting given that they cannot utilize the common knowledge shared across different graph datasets.

\begin{figure}[tb!]
  \centering
  \includegraphics[width=\linewidth]{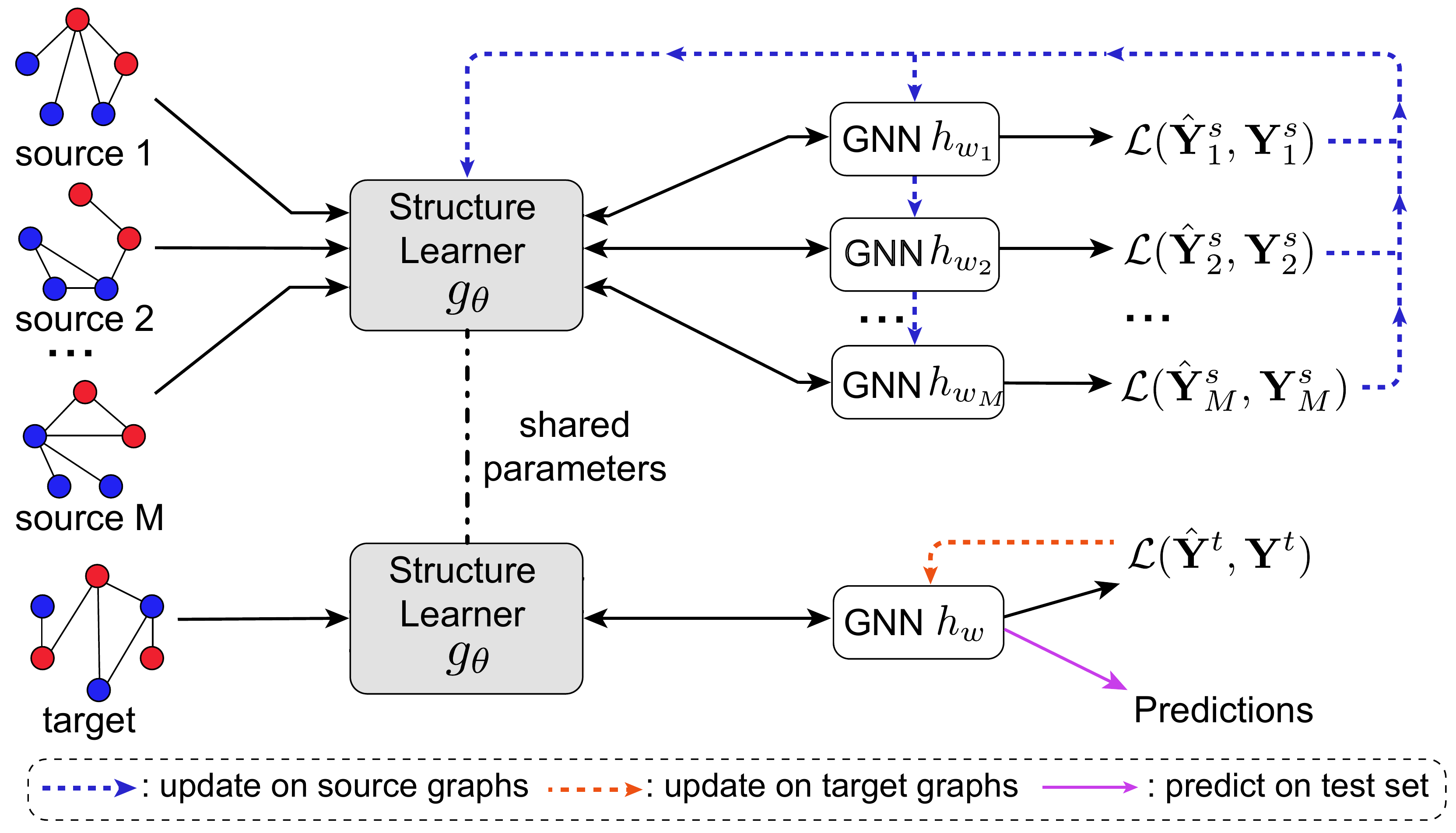}
  \caption{Illustration of Open-World Graph Structure Learning. In a diverse set of source graphs, we train multiple dataset-specific GNNs and a shared structure learner. In the target graph, we directly utilize the learned structure learner and only need to train a new GNN.}
  \label{fig:glow}
\end{figure}

To resolve the above dilemma, this paper attempts to explore a novel problem setting termed \textit{Open-World Graph Structure Learning}. Specifically, we target learning a generalizable graph structure learning model which is trained with multiple source graphs and can be directly adapted for inference (without re-training or fine-tuning) on new unseen target graphs. We formulate the problem as a bi-level optimization target that jointly learns a single dataset-shared structure learner and multiple dataset-specific GNNs tailored for particular graph datasets, as shown in Fig. \ref{fig:glow}. Under such a framework, the well-trained structure learner can leverage the common transferrable knowledge across datasets for enhancing generalization and more critically, be readily utilized to yield adaptive message-passing topology for arbitrarily given target graphs.

With the guidance of the aforementioned general goal, we propose \mymodel (short for A Graph Structure Learning Model for Open-World Generalization) that aims at learning the generalizable patterns of optimal message-passing topology across source graphs. Specifically, we first take a bottom-up perspective and formulate the generative process for observed data in a probabilistic manner. On top of this, we derive a tractable and feasible learning objective through the lens of variational inference. The structure learner is specified as a multi-head weighted similarity function so as to guarantee enough expressivity for accommodating diverse structural information, and we further harness an approximation scheme to reduce the quadratic complexity overhead of learning potential edges from arbitrary node pairs. 

To reasonably and comprehensively evaluate the model, we devise experiments with a diverse set of protocols that can measure the generalization ability under different difficulty levels (according to the intensity of distribution shifts between source graphs and target graphs).
Concretely, we consider: 1) In-domain generalization, in which we generalize from some citation (social) networks to other citation (social) networks. 2) Cross-domain networks generalization between citation and social networks.
The results, which are consistent across various combinations of source and target graph datasets, demonstrate that when evaluated on the target graphs, our approach i) consistently outperforms directly training the GNN counterpart on original non-optimized graph structures of the target datasets and ii) performs on par with state-of-the-art structure learning methods~\cite{lds,idgl,variational_inference_2} trained on target graphs from scratch with up to $25\times$ less training time consumed. 
Our code is available at \href{https://github.com/WtaoZhao/GraphGLOW}{https://github.com/WtaoZhao/GraphGLOW}.

	\begin{figure*}[t!]
		\centering
		\includegraphics[width=0.86\linewidth]{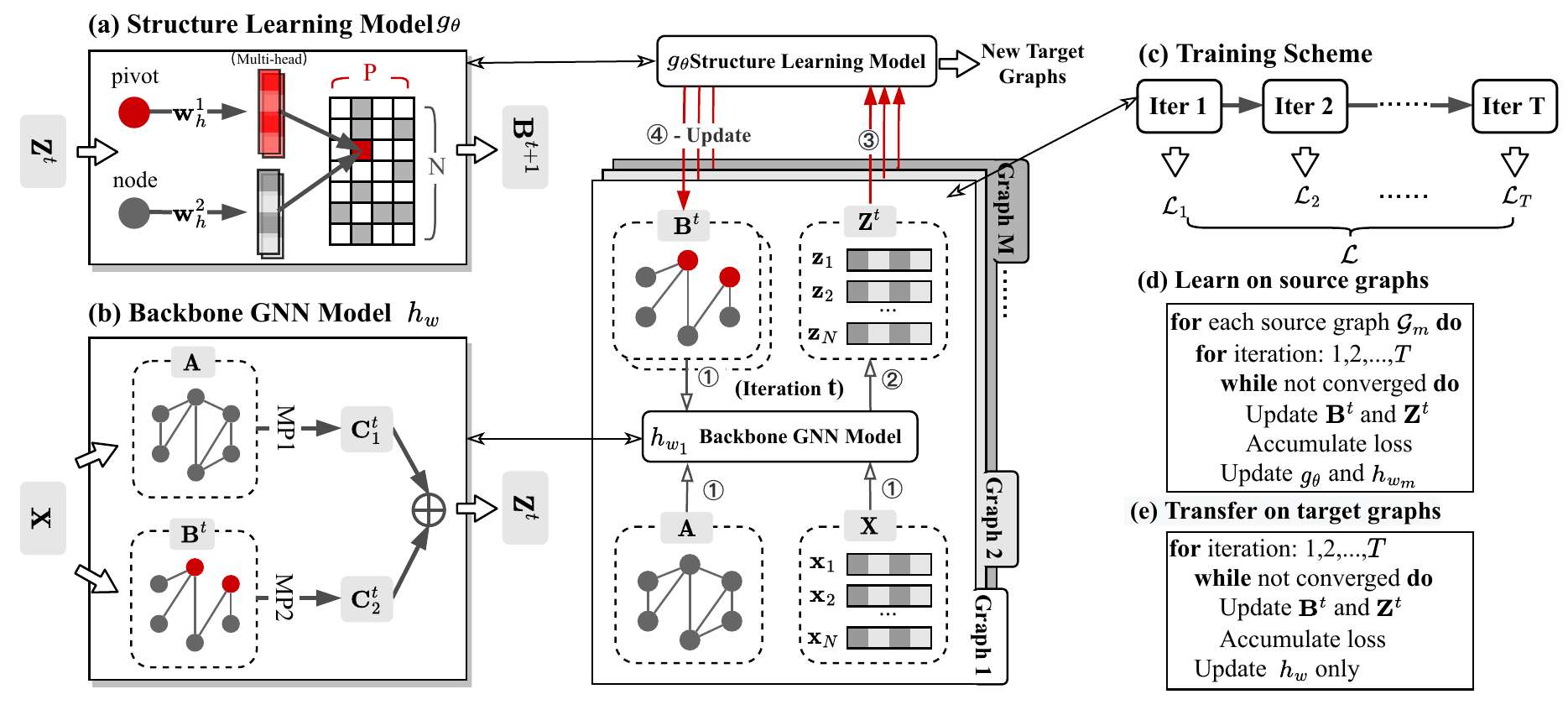}
		\caption{Illustration of the proposed framework \mymodel targeting open-world graph structure learning. The middle part of the figure presents the training process for the structure learner together with multiple dataset-specific GNNs on source graphs. In (a)-(e) we illustrate the details of graph structure learner, backbone GNN, iterative training process, training procedure and transferring procedure.
		When the training is finished, the structure learner is fixed and we only need to train a dataset-specific GNN network on new target graph with latent structures inferred by the well-trained structure learner.
		} 
		\label{fig:precis}
	\end{figure*}

\section{Preliminary and Problem Definition}\label{sec-problem}


\textbf{Node-Level Predictive Tasks.} Denote a graph with $N$ nodes as $\mathcal G = (\mathbf A, \mathbf X, \mathbf Y)$ where $\mathbf A = \{a_{uv}\}_{N\times N}$ is an adjacency matrix ($a_{uv}=1$ means the edge between node $u$ and $v$ exists and 0 otherwise), $\mathbf X = \{\mathbf x_u\}_{N\times D}$ is a feature matrix with $\mathbf x_u$ a $D$-dimensional node feature vector of node $u$, and $\mathbf Y = \{y_u\}_{N\times C}$ with $y_u$ the label vector of node $u$ and $C$ class number. The node labels are partially observed as training data, based on which the node-level prediction aims to predict the unobserved labels for testing nodes in the graph using node features and graph structures. The latter is often achieved via a GNN model, denoted as $h_w$, that yields predicted node labels $\hat{\mathbf Y} = h_w(\mathbf A, \mathbf X)$ and is optimized with the classification loss $w^* = \argmin_{w} = \mathcal L(\hat{\mathbf Y}, \mathbf Y)$ using observed labels from training nodes.

\textbf{Closed-World Graph Structure Learning (GLCW).}
 The standard graph structure learning for node-level predictive tasks trains a graph structure learner $ g_\theta $ to refine the given structure, i.e., $ \hat{\mathbf{A}}=g_\theta (\mathbf{A}, \mathbf{X}) $, over which the GNN classifier $ h_w $ conducts message passing for producing node representations and predictions. The $g_\theta$ is expected to produce optimal graph structures that can give rise to satisfactory downstream classification performance of the GNN classifier. Formally speaking, the goal for training $ g_\theta $ along with $ h_w $ can be expressed as a nested optimization problem:
\begin{equation}\label{eqn-standard}
	\theta^* = \argmin_{w}  \min_{\theta}   \mathcal L \left (h_{w}(g_\theta(\mathbf A, \mathbf X), \mathbf X),\mathbf Y \right ).
\end{equation}
The above formulation of graph structure learning under closed-world assumptions constrains the training and testing nodes in the same graph, which requires $g_\theta$ to be trained from scratch on each graph dataset. Since $g_\theta$ is often much more complicated (e.g., with orders-of-magnitude more trainable parameters) and difficult for optimization (due to the bi-level optimization \eqref{eqn-standard}) than the GNN $h_w$, the GLCW would lead to undesired inefficiency and vulnerability for serious over-fitting (due to limited labeled information). 

\textbf{Open-World Graph Structure Learning (GLOW).}
In this work, we turn to a new learning paradigm that generalizes graph structure learning to open-world assumptions, borrowing the concepts of domain generalization~\cite{dg-1} and out-of-distribution generalization~\cite{wu2022handling}, more broadly. Specifically, assume that we are given multiple source graphs, denoted as $\{\mathcal G^s_m\}_{m=1}^M = \{(\mathbf A^s_m, \mathbf X^s_m, \mathbf Y^s_m)\}_{m=1}^M$, and a target graph $\mathcal G^t = (\mathbf A^t, \mathbf X^t, \mathbf Y^t)$, whose distribution is often different from any source graph. The goal is to train a universal structure learner $ g_\theta $ on source graphs which can be directly used for inference on the target graph without any re-training or fine-tuning. The trained structure learner is expected to produce desired graph structures that can bring up better downstream classification of a GNN classifier optimized for the target graph.

More specifically, we consider a one-to-many framework that coordinates a shared graph structure learner $ g_\theta $ and multiple dataset-specific GNNs $ \{h_{w_m} \}_{m=1}^M$, where $h_{w_m}$ with independent parameterization $w_m$ is optimized for a given source graph $ \mathcal{G}_m^s $. With the aim of learning a universal $ g_\theta $ that can generalize to new unseen target graphs, our training goal can be formulated as the following bi-level optimization problem:
\begin{equation}\label{eqn-obj}
	\theta^* = \argmin_{\theta}  \min_{w_1, \cdots, w_M} \sum_{m=1}^M  \mathcal L \left (h_{w_m}(g_\theta(\mathbf A^s_m, \mathbf X^s_m), \mathbf X^s_m),\mathbf Y^s_m \right ),
\end{equation}
where the inner optimization is a multi-task learning objective. Generally, \eqref{eqn-obj} aims at finding an optimal $g_\theta$ that can jointly minimize the classification loss induced by $M$ GNN models, each trained for a particular source graph.
After training, we can directly adapt $g_{\theta^*}$ to the target graph for testing purpose, and only need to train a GNN $h_w$ on the target graph:
\begin{equation}
	w^* = \argmin_{w} \mathcal L \left (h_{w}(g_{\theta^*}(\mathbf A^t, \mathbf X^t), \mathbf X^t),\mathbf Y^t \right ).
\end{equation}

\section{Proposed Model}

To handle the above problem, we present an end-to-end learning framework \mymodel that guides the central graph structure learner to learn adaptive message-passing structures exploited by multiple GNNs. The overview of \mymodel is shown in Fig.~\ref{fig:precis}.

The fundamental challenge of GLOW lies in how to model and capture the generalizable patterns among adaptive structures of different graphs. To this end, we first take a data-generative perspective that treats the inputs and inter-mediate results as random variables and investigate into their dependency, based on which we present the high-level model formulation in a probabilistic form (Sec.~\ref{sec-model-for}). Then we proceed to instantiate the model components (Sec.~\ref{sec-model-ins}). Finally, we discuss differentiable training approaches for optimization (Sec.~\ref{sec-model-training}).

\subsection{Model Formulation}\label{sec-model-for}

To commence, we characterize the data generation process by a latent variable model, based on which we derive the formulation of our method. We treat the latent graph $\hat{\mathbf A}$ (given by $g_\theta$) as a latent variable whose prior distribution is given by $p(\hat{\mathbf A} | \mathbf A, \mathbf X)$. The prior distribution reflects how one presumed on the latent structures before observed labels arrive. Then, the prediction is given by a predictive distribution $p(\mathbf Y| \hat{\mathbf A}, \mathbf A, \mathbf X)$. 
The learning objective aims at maximizing the log-likelihood of observed labels, which can be written as:
	$ \log p(\mathbf{Y} | \mathbf{A} , \mathbf{X}  )
	=\log \int_{\hat{\mathbf{A} }} p(\mathbf{Y} | \mathbf{A}, \mathbf{X}, \hat{\mathbf{A} }   ) p(\hat{\mathbf{A} } | \mathbf{A}, \mathbf{X}   ) d \hat{\mathbf{A} } $.
To estimate latent graphs that could enhance message passing for downstream tasks, one plausible way is to sample from the posterior, i.e., $ p(\hat{\mathbf{A} } | \mathbf Y, \mathbf{A}, \mathbf{X}) $,  conditioned on the labels from downstream tasks. Using the Bayes' rule, we have
\begin{equation}
    p(\hat{\mathbf{A} } | \mathbf Y, \mathbf{A}, \mathbf{X}) = \frac{ p(\mathbf{Y} | \mathbf{A}, \mathbf{X}, \hat{\mathbf{A} }   ) p(\hat{\mathbf{A} } | \mathbf{A}, \mathbf{X}   )}{\int_{\hat{\mathbf{A} }} p(\mathbf{Y} | \mathbf{A}, \mathbf{X}, \hat{\mathbf{A} }   ) p(\hat{\mathbf{A} } | \mathbf{A}, \mathbf{X}   ) d \hat{\mathbf{A} }}.
\end{equation}
However, the integration over $\hat{\mathbf A}$ in the denominator is intractable  for computation due to the exponentially large space of $\hat{\mathbf{A}}$. 

To circumvent the difficulty, we can introduce a variational distribution $ q(\hat{\mathbf{A} } | \mathbf{A}, \mathbf{X}   ) $ over $ \hat{\mathbf{A} } $ as an approximation to $p(\hat{\mathbf{A} } | \mathbf Y, \mathbf{A}, \mathbf{X})$. We can sample latent graphs from $ q(\hat{\mathbf{A} } | \mathbf{A}, \mathbf{X}   ) $, i.e., instantiate it as the structure learner $g_\theta$, and once $q(\hat{\mathbf{A} } | \mathbf{A}, \mathbf{X}   ) = p(\hat{\mathbf{A} } | \mathbf Y, \mathbf{A}, \mathbf{X})$, we could have samples from the posterior that ideally generates the optimal graph structures for downstream prediction. By this principle, we can start with minimizing the Kullback-Leibler divergence between $q$ and $p$ and derive the learning objective as follows:
\begin{equation}
	\begin{split}
		& \mathcal D_{KL}(q(\hat{\mathbf{A} } | \mathbf{A}, \mathbf{X}   ) \| p(\hat{\mathbf{A} } | \mathbf Y, \mathbf{A}, \mathbf{X})) \\
		= & - \underbrace{\mathbb{E}_{\hat{\mathbf{A} } \sim q(\hat{\mathbf{A} } | \mathbf{A}, \mathbf{X}   )} \left [\log  \frac{ p(\mathbf{Y} | \mathbf{A}, \mathbf{X}, \hat{\mathbf{A} }   )  p(\hat{\mathbf{A} } | \mathbf{A}, \mathbf{X}   ) }{ q(\hat{\mathbf{A} } | \mathbf{A}, \mathbf{X}   ) } \right ] }_{\mbox{Evidence Lower Bound}} + \log p(\mathbf Y|\mathbf A, \mathbf X).
	\end{split}
\end{equation}
Based on this equation, we further have the inequality which bridges the relationship between the Evidence Lower Bound (ELBO) and observed data log-likelihood:
\begin{equation}
    \log p(\mathbf Y|\mathbf A, \mathbf X) \\
    \geq  \mathbb{E}_{\hat{\mathbf{A} } \sim q(\hat{\mathbf{A} } | \mathbf{A}, \mathbf{X}   )} \left [\log  \frac{ p(\mathbf{Y} | \mathbf{A}, \mathbf{X}, \hat{\mathbf{A} }   )  p(\hat{\mathbf{A} } | \mathbf{A}, \mathbf{X}   ) }{ q(\hat{\mathbf{A} } | \mathbf{A}, \mathbf{X}   ) } \right ].
\end{equation}
The equality holds if and only if $\mathcal D_{KL}(q(\hat{\mathbf{A} } | \mathbf{A}, \mathbf{X}) \| p(\hat{\mathbf{A} } | \mathbf Y, \mathbf{A}, \mathbf{X})) = 0$. The above fact suggests that we can optimize the ELBO as a surrogate for $\log p(\mathbf Y|\mathbf A, \mathbf X)$ which involves the intractable integration.
More importantly, when the ELBO is optimized w.r.t. $q$ distribution, the variational bound is lifted to the original log-likelihood and one has $q(\hat{\mathbf{A} } | \mathbf{A}, \mathbf{X})=p(\hat{\mathbf{A} } | \mathbf Y, \mathbf{A}, \mathbf{X})$, i.e., the variational distribution equals to the true posterior, which is what we expect.  

Pushing further and incorporating source graphs $\mathcal G_m$ (we omit the superscript for simplicity), we arrive at the following objective:
\begin{equation}\label{eqn-elbo}
\begin{split}
    \mathbb E_{\mathcal G_m \sim p(\mathcal G)} \left [ 
    \mathbb{E}_{\hat{\mathbf{A} } \sim q_\theta(\hat{\mathbf{A} } | \mathbf{A} = \mathbf A_m, \mathbf{X} = \mathbf X_m   )} \left [\log p_{w_m }(\mathbf{Y} | \mathbf{A} = \mathbf A_m, \mathbf{X} = \mathbf X_m, \hat{\mathbf{A} }  ) \right. \right. \\
    \left. \left. + \log p_{0}(\hat{\mathbf{A} } | \mathbf{A} = \mathbf A_m, \mathbf{X} = \mathbf X_m   ) - \log q_\theta(\hat{\mathbf{A} } | \mathbf{A} = \mathbf A_m, \mathbf{X} = \mathbf X_m   )  \right ]
    \right ].
\end{split}
\end{equation}
Here we instantiate $q(\hat{\mathbf{A} } | \mathbf{A}, \mathbf{X}   )$ as the shared structure learner $g_\theta$, $p(\hat{\mathbf A}|\mathbf A, \mathbf X)$ as a (shared) non-parametric prior distribution $p_0$ for latent structures, and $p(\mathbf{Y} | \mathbf{A}, \mathbf{X}, \hat{\mathbf{A}})$ as the dataset-specific GNN model $h_{w_m}$, to suit the framework for our formulated problem in Section~\ref{sec-problem}. The formulation of \eqref{eqn-elbo} shares the spirits with Bayesian meta learning~\cite{maml-bayes}. We can treat the GNN training as a dataset-specific learning task and latent graph as a certain `learning algorithm' or `hyper-parameter', so \eqref{eqn-elbo} essentially aims at learning a structure learner that can yield desirable `learning algorithm' for each specific learning task on graphs. Furthermore, the three terms in \eqref{eqn-elbo} have distinct effects: i) the predictive term $\log p_{w_m}$ acts as a supervised classification loss; ii) the prior term $\log p_0$ serves for regularization on the generated structures; iii) the third term, which is essentially the entropy of $q_\theta$, penalizes high confidence on certain structures. 

To sum up, we can optimize \eqref{eqn-elbo} with joint learning of the structure learner $g_\theta$ and GNN models $\{h_{w_m}\}_{m=1}^M$ on source graphs $\{\mathcal G_m\}_{m=1}^M$ for training the structure learner. After that, we can generalize the well-trained $g_{\theta^*}$ to estimate latent graph structures for a new target graph $\mathcal G^t = (\mathbf A^t, \mathbf X^t)$ and only need to train the GNN model $h_w$ w.r.t. the predictive objective with fixed $\theta^*$:
\begin{equation}
    \mathbb{E}_{\hat{\mathbf{A} } \sim q_{\theta^*}(\hat{\mathbf{A} } | \mathbf{A} = \mathbf A^t, \mathbf{X} = \mathbf X^t   )} \left [\log p_{w}(\mathbf{Y} | \mathbf{A} = \mathbf A^t, \mathbf{X} = \mathbf X^t, \hat{\mathbf{A} }   ) \right ].
\end{equation}

We next discuss how to specify $g_\theta$, $h_{w_m}$ and $p_0$ with special focus on their expressiveness and efficiency in Section~\ref{sec-model-ins}. Later, we present the details for loss computation and model training based on the formulation stated above in Section~\ref{sec-model-training}.

\subsection{Model Instantiations}\label{sec-model-ins}

\subsubsection{Instantiation for $q_\theta(\hat{\mathbf A}|\mathbf A, \mathbf X)$}

The variational distribution aims at learning the conditional distribution that generates suitable latent structures for message passing based on input observations. A natural means is to assume each edge of the latent graph as a Bernoulli random variable and the distribution $q$ is a product of $N\times N$ independent Bernoulli random variables~\cite{variational_inference_2,GIB-neurips20}.

The graph structure learner $g_\theta$ can be used for predicting the Bernoulli parameter matrix. To accommodate the information from node features and graph structure, we can use the node representation, denoted as $\mathbf z_u \in \mathbb R^d$, where $d$ is the embedding dimension, to compute the edge probability $\alpha_{uv}$ for edge $(u,v)$ as
\begin{equation}\label{eqn-sim}
    \alpha_{uv}= \delta\left (\frac{1}{H}\sum_{h=1}^H s(\mathbf {w}_{h}^1 \odot \mathbf {z}_u, \mathbf {w}_{h}^2 \odot \mathbf {z}_v) \right ),
\end{equation}
where $s(\cdot, \cdot)$ is a similarity function for two vectors, $\odot$ denotes Hadamard product, $\delta$ is a function that converts the input into values within $[0,1]$ and $\mathbf {w}_{h}^1, \mathbf {w}_{h}^2 \in \mathbb R^d$ are two weight vectors of the $h$-th head. Common choices for $s(\cdot, \cdot)$ include simple dot-product, cosine distance \cite{cosine}, RBF kernel \cite{rbf}, etc. Here we introduce $H$ heads and aggregate their results to enhance model's expressiveness for capturing the underlying influence between nodes from multifaceted causes, following the spirit of multi-head attention~\cite{transformer,gat}. Besides, the weight vectors in \eqref{eqn-sim} could learn to element-wisely scale the input vectors, i.e., node representations, and adaptively attend to dominant features. Apart from these, two weight vectors $\mathbf {w}_{h}^1, \mathbf {w}_{h}^2$ with independent parameterization could potentially have the same or distinct directions, which makes the model capable of connecting similar or dissimilar nodes and expressive enough to handle both homophilous and non-homophilous graphs. 

\begin{figure}[t!]
	\centering
	\includegraphics[width=0.47\textwidth]{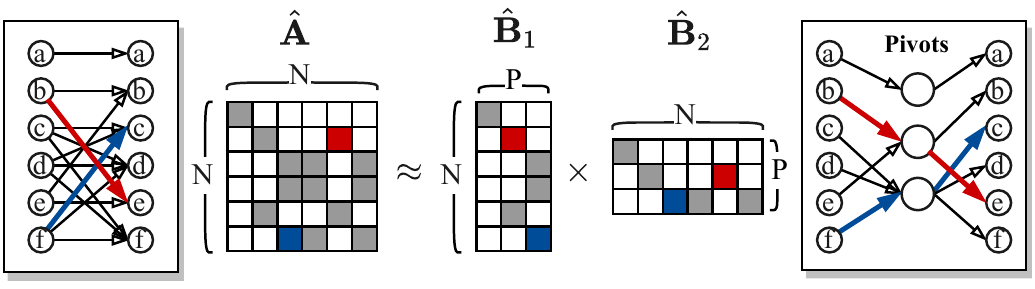}
	\caption{Illustration for scalable structure learning  message passing, which reduces algorithmic complexity from $O(N^2)$ to $O(NP)$. We choose $P$ nodes as pivots and convert the $N\times N$ matrix  to the product of two $N\times P$ node-pivot matrices (where the message passing is executed with two steps, i.e., node-to-pivot and pivot-to-node.). \label{fig-pivot}}
\end{figure}

To obtain discrete latent graph $\hat{\mathbf A} = \{\hat a_{uv}\}_{N\times N}$, one can sample from $\hat a_{uv} \sim Bernoulli(\alpha_{uv})$ to obtain each latent edge. 
However, such an approach induces the quadratic algorithmic complexity $O(N^2)$ for computing and storing an estimated structure that entails potential links between any node pair, which could be prohibitive for large graphs. To reduce space and time complexity, we adopt a pivot-based structure learning method, as shown in Fig.~\ref{fig-pivot}. Concretely, we randomly choose $P$ nodes in the graph as \textit{pivots}, where
$P$ is a hyperparameter much smaller than $N$ (e.g., $P\approx \frac{1}{10}N$). We then leverage pivot nodes as intermediates and convert the $N\times N$ graph $\hat {\mathbf A}$, which can be prohibitively large with dense edges, into a cascade of one $N\times P$ node-pivot bipartite graph $\hat{\mathbf B}_1$ and one $P\times N$ pivot-node bipartite graph $\hat{\mathbf B}_2 = \hat{\mathbf B}_1^\top$, which can effectively control the computational cost with proper $P$. In this way, we can compute a node-pivot similarity matrix $\mathbf \Gamma = \{\alpha_{up}\}_{N\times P}$ based on \eqref{eqn-sim}, to parameterize the distribution of latent graph structures. This only requires $O(NP)$ time and space complexity, and one can sample from each $Bernoulli(\alpha_{up})$ to obtain $\hat{\mathbf B}_1$ and $\hat{\mathbf B}_2$. In the meanwhile, the original $N\times N$ adjacency matrix could be retrieved by $\hat{\mathbf A} = \hat{\mathbf B}_1 \hat{\mathbf B}_2$, which suggests that one can execute message passing on $\hat{\mathbf B}_1$ and $\hat{\mathbf B}_2$ to approximate that on $\hat{\mathbf A}$ (see more details in Section~\ref{sec-method-ins-p}). In terms of the acceleration of structure learning, other strategies like the all-pair message passing schemes with linear complexity explored by \cite{wunodeformer,wudifformer} can also be utilized to achieve the purpose.


\subsubsection{Instantiation for $p_{w_m}(\mathbf Y|\mathbf A, \mathbf X, \hat{\mathbf A})$}\label{sec-method-ins-p}

The predictive distribution, parameterized by the GNN network $h_{w_m}$, aims at recursively propagating features along the latent graph to update node representations and producing the prediction result for each node. We then present the details of GNN's message passing on the latent graph in order for enough expressiveness, stability and efficiency.

To begin with, we review the message-passing rule in common GNN models, like GCN~\cite{gcn}, that operates on the original graph $\mathbf A$:
\begin{equation}\label{eqn-mp-gcn}
	\mathbf Z^{(l+1)} = \sigma \left (\operatorname{MP}_1(\mathbf{Z}^{(l)},\mathbf{A}) \mathbf W^{(l)} \right )= \sigma \left( \mathbf D^{-\frac{1}{2}} \mathbf A \mathbf D^{-\frac{1}{2}} \mathbf{Z}^{(l)} \mathbf W^{(l)} \right ), 
\end{equation}
where $\mathbf W^{(l)}\in \mathbb R^{d\times d}$ is a weight matrix, $\sigma$ is non-linear activation, and $\mathbf D$ denotes a diagonal degree matrix from input graph $\mathbf A$ and $\mathbf Z^{(l)} = \{\mathbf z_u^{(l)}\}_{N\times d}$ is a stack of node representations at the $l$-th layer. 

With the estimated latent graph $\hat{\mathbf A} = \hat{\mathbf B}_1 \hat{\mathbf B}_2$, we perform message passing $\mathrm{MP}_2(\cdot)$ in a two-step fashion to update node representations:
\begin{equation}
    \begin{split}
        & \mbox{i) node-to-pivot passing:} ~~~ \mathbf C^{(l+\frac{1}{2})} = \mathrm{RowNorm} (\mathbf \Gamma^\top) \mathbf Z^{(l)}, \\
        & \mbox{ii) pivot-to-node passing:} ~~~ \mathbf C^{(l+1)} = \mathrm{RowNorm} ( \mathbf \Gamma) \mathbf C^{(l+\frac{1}{2})},
    \end{split}
\end{equation}
where $\mathbf C^{(l+\frac{1}{2})}$ is an intermediate node representation and $\mathbf \Gamma = \{\alpha_{uv}\}_{N\times P}$ is the node-pivot similarity matrix calculated by \eqref{eqn-sim}.  Such a two-step procedure can be efficiently conducted within $O(NP)$ time and space complexity. 


Despite that the feature propagation on the estimated latent structure could presumably yield better node representations, the original input graph structures also contain useful information, such as effective inductive bias~\cite{geometriclearning-2016}. Therefore, we integrate two message-passing functions to compute layer-wise updating for node representations:
\begin{equation}
    \mathbf Z^{(l+1)} = \sigma \left( \lambda \mbox{MP}_1(\mathbf Z^{(l)}, \mathbf A) \mathbf W^{(l)} + (1 - \lambda) \mbox{MP}_2(\mathbf Z^{(l)}, \hat{\mathbf A}) \mathbf W^{(l)} \right ),
\end{equation}
where $\lambda$ is a trading hyper-parameter that controls the concentration weight on input structures. Such design could also improve the training stability by reducing the impact from large variation of latent structures through training procedure. 

With $L$ GNN layers, one can obtain the prediction $\hat {\mathbf Y}$ by setting $\hat {\mathbf Y} = \mathbf Z^{(L)}$ and $\mathbf W^{(L-1)} \in \mathbb R^{d\times C}$ where $C$ is the number of classes. Alg.~\ref{alg:GNN} shows the feed-forward computation of message passing.

\subsubsection{Instantiation for $p_0(\hat{\mathbf A}|\mathbf A, \mathbf X)$}

The prior distribution reflects how we presume on the latent graph structures without the information of observed labels. In other words, it characterizes how likely a given graph structure could provide enough potential for feature propagation by GNNs. The prior could be leveraged for regularization on the estimated latent graph $\hat{\mathbf A}$. In this consideration, we choose the prior as an energy function that quantifies the smoothness of the graph:
\begin{equation}\label{eqn-reg}
    p_0(\hat{\mathbf A}| \mathbf X, \mathbf A) \propto \exp{\left( -\alpha \sum_{u, v} \hat{\mathbf A}_{uv} \| \mathbf x_u - \mathbf x_v \|_2^2 
    - \rho \|\hat{\mathbf A}\|_F^2 \right ) },
\end{equation}
where 
$\| \cdot \|_{F}$ is the Frobenius norm. The first term in \eqref{eqn-reg} measures the smoothness of the latent graph~\cite{dirichlet}, with the hypothesis that graphs with smoother feature has lower energy (i.e., higher probability).
The 
second term helps avoiding too large node degrees~\cite{connectivity}. The hyperparameters
$ \alpha $ and $ \rho $ control the strength for regularization effects.

While we can retrieve the latent graph via $\hat{\mathbf A} = \hat{\mathbf B}_1 \hat{\mathbf B}_2$, the computation of \eqref{eqn-reg} still requires $ O(N^2) $ cost. To reduce the overhead, we apply the regularization on the $ P \times P $ pivot-pivot adjacency matrix $\hat{\mathbf{E}} = \hat{\mathbf B}_2\hat{\mathbf B}_1$ as a proxy regularization: 
\begin{equation}\label{eqn-reg1}
\begin{split}
    \mathcal R (\hat{\mathbf E}) &= \log p_0(\hat{\mathbf A}| \mathbf X, \mathbf A) \\
    & \approx -\alpha \sum_{p, q} \hat{\mathbf E}_{pq} \| \mathbf x'_p - \mathbf x'_q \|_2^2 - \rho \|\hat{\mathbf E}\|_F^2,
\end{split}
\end{equation}
where $\mathbf x'_p$ denotes the input feature of the $p$-th pivot node.

\subsection{Model Training}\label{sec-model-training}

For optimization with \eqref{eqn-elbo}, we proceed to derive the loss functions and updating gradients for $\theta$ and $w_m$ based on the three terms $\mathbb E_{q_\theta}[\log p_{w_m}]$, $\mathbb E_{q_\theta}[\log p_0]$ and $\mathbb E_{q_\theta}[\log q_\theta]$. 

\subsubsection{Optimization for $\mathbb E_{q_\theta}[\log p_{w_m}]$}\label{sec-model-training-p}
The optimization difficulty stems from the expectation over $q_\theta$, where the sampling process is non-differentiable and hinders back-propagation. 
Common strategies for approximating the sampling for discrete random variables include Gumbel-Softmax trick~\cite{gumbel-iclr17} and REINFORCE trick~\cite{reinforce}. However, both strategies yield a sparse graph structure each time of sampling, which could lead to high variance for the prediction result $\log p_{w_m}(\mathbf Y| \mathbf A, \mathbf X, \hat{\mathbf A})$ produced by message passing over a sampled graph. To mitigate the issue, we alternatively adopt the Normalized Weighted Geometric Mean (NWGM)~\cite{NWGM} to move the outer expectation into the feature-level. Specifically, we have (see Appendix~\ref{appx-deri} for detailed derivations)
\begin{equation}\label{eqn-loss-sup-grad}
    \begin{split}
        & \nabla_{\theta} \mathbb E_{q_\theta(\hat{\mathbf A}|\mathbf A, \mathbf X)} [ \log p_{w_m} (\mathbf Y| \mathbf A, \mathbf X, \hat{\mathbf A})] \\ 
        \approx & \nabla_{\theta}  \log p_{w_m} (\mathbf Y| \mathbf A, \mathbf X, \hat{\mathbf A} = \mathbb E_{q_\theta(\hat{\mathbf A}|\mathbf A, \mathbf X)} [\hat{\mathbf A}]).
    \end{split}
\end{equation}
We denote the opposite of the above term  as $ \nabla_\theta \mathcal{L}_s(\theta) $.  The gradient w.r.t. $w_m$ can be similarly derived. The above form is a biased estimation for the original objective, yet it can reduce the variance from sampling and also improve training efficiency (without the need of message passing over multiple sampled graphs).\eqref{eqn-loss-sup-grad} induces the supervised cross-entropy loss.

\subsubsection{Optimization for $\mathbb E_{q_\theta}[\log p_0]$}
As for the second term in \eqref{eqn-elbo}, we adopt the REINFORCE trick, i.e., policy gradient, to tackle the non-differentiability of sampling from $q_\theta$. Specifically, for each feedforward computation, we sample from the Bernoulli distribution for each edge given by the estimated node-pivot similarity matrix, i.e., $Bernoulli(\alpha_{up})$, and obtain the sampled latent bipartite graph $\hat{\mathbf B}_1$ and subsequently have $\hat{\mathbf E} = \hat{\mathbf B}_1 \hat{\mathbf B}_2 = \hat{\mathbf B}_1 \hat{\mathbf B}_1^\top$. The probability for the latent structure could be computed by 
\begin{equation}\label{eqn-pi}
\pi_\theta(\hat{\mathbf E}) = \prod_{u, p} \left( \hat{\mathbf B}_{1,up}  \alpha_{up} + (1 - \hat{\mathbf B}_{1,up}) \cdot(1 - \alpha_{up})    \right).
\end{equation}
Denote $\hat{\mathbf E}_k$ as the sampled result at the $k$-th time, we can independently sample $K$ times and obtain $\{\hat{\mathbf E}_k\}_{k=1}^K$ and $\{\pi_\theta(\hat{\mathbf E}_k)\}_{k=1}^K$. Recall that the regularization reward from $\log p_0$ has been given by \eqref{eqn-reg1}. 
The policy gradient~\cite{reinforce} yields the gradient of loss for $\theta$ as
\begin{equation}\label{eqn-grad-reg}
\begin{split}
    \nabla_{\theta} \mathcal L_r(\theta) &= -\nabla_{\theta} \mathbb E_{\hat{\mathbf{A}}\sim q(\hat{\mathbf{A}}  |  \mathbf{X},\mathbf{A} )} [\log p_0(\hat{\mathbf{A}}  |  \mathbf{X},\mathbf{A})] \\
    &\approx -\nabla_{\theta} \frac{1}{K} \sum_{k=1}^K \log \pi_\theta(\hat {\mathbf E}_k) \left(\mathcal R(\hat{\mathbf E}_k) - \overline{\mathcal R}\right),
\end{split}
\end{equation}
where $\overline{\mathcal R}$ acts as a baseline function by averaging the regularization rewards $\mathcal R(\hat{\mathbf E}_k)$ in one feed-forward computation, which helps to reduce the variance during policy gradient training~\cite{policy}. 

\subsubsection{Optimization with $\mathbb E_{q_\theta}[\log q_\theta]$}

The last entropy term for $q_\theta$ could be directly computed by
\begin{equation}\label{eqn-loss-ent}
\begin{split}
    \mathcal L_{e} (\theta) &= \mathbb E_{\hat{\mathbf{A}}\sim q(\hat{\mathbf{A}}  |  \mathbf{X},\mathbf{A} )} [\log q(\hat{\mathbf{A}}  |  \mathbf{X},\mathbf{A} )] \\
    &\approx
		\frac{1}{NP} \sum_{u=1}^{N} \sum_{p=1}^{P} \left[ \alpha_{up}\log \alpha_{up}  + (1- \alpha_{up}  ) \log (1-\alpha_{up} )  \right],
\end{split}
\end{equation}
where we again adopt the node-pivot similarity matrix as a proxy for the estimated latent graph.

\begin{table*}[tb!]
	\caption{Test accuracy (\%) on target graphs for in-domain generalizations. For each social network (resp. citation network) as target dataset, we consider the other social networks (resp. citation networks) as source graphs. 
		\mymodelvar is an oracle model that shares the same architecture as our model \mymodel and is directly trained on target graphs. \label{tab-res1}}
		\begin{tabular}{| c| lccccccc |}
			\hline
			\textbf{Type} & \textbf{Method }& \textbf{Cornell5} & \textbf{Johns.55} & \textbf{Amherst41} & \textbf{Reed98} & \textbf{Cora} & \textbf{CiteSeer} & \textbf{PubMed} \\
			\hline
			\multirow{7}{*}{\specialcell[t]{\textbf{Pure}\\\textbf{GNN}} } & GCN & 68.6 ± 0.5 & 70.8 ± 1.0 & 65.8 ± 1.6 & 60.8 ± 1.6 & 81.6 ± 0.4 & 71.6 ± 0.3 & 78.8 ± 0.6 \\
			& SAGE & 68.7 ± 0.8 & 67.5 ± 0.9 & 66.3 ± 1.8 & 63.9 ± 1.9 & 81.4 ± 0.6 & 71.6 ± 0.5 & 78.6 ± 0.7 \\
			& GAT & 69.6 ± 1.2 & 69.4 ± 0.7 & 68.7 ± 2.1 & 64.5 ± 2.5 & 83.0 ± 0.7 & 72.1 ± 1.1 & 79.0 ± 0.4 \\
			& GPR & 68.8 ± 0.7 & 69.6 ± 1.3 & 66.2 ± 1.5 & 62.7 ± 2.0 & 83.1 ± 0.7 & 72.4 ± 0.8 & 79.6 ± 0.5 \\
			& APPNP & 68.5 ± 0.8 & 69.1 ± 1.4 & 65.9 ± 1.3 & 62.3 ± 1.5 & 82.7 ± 0.5 & 71.9 ± 0.5 & 79.2 ± 0.3 \\
			& H$_2$GCN & 71.4 ± 0.5 & 68.3 ± 1.0 & 66.5 ± 2.2 & 65.4 ± 1.3 & 82.5 ± 0.8 & 71.4 ± 0.7 & 79.4 ± 0.4 \\
			& CPGNN & 71.1 ± 0.5 & 68.7 ± 1.3 & 66.7 ± 0.8 & 63.6 ± 1.8 & 80.8 ± 0.4 & 71.6 ± 0.4 & 78.5 ± 0.7 \\
			\hline
			\multirow{6}{*}{\specialcell[t]{\textbf{Graph}\\\textbf{Structure}\\\textbf{Learning}}}  & $\mathrm{\mymodel_{dp}}$ & 71.5 ± 0.7 & 71.3 ± 1.2 & 68.5 ± 1.6 & 63.2 ± 1.2 & 83.1 ± 0.8 & 71.7 ± 1.0 & 77.3 ± 0.8 \\
			& $\mathrm{\mymodel_{knn}}$& 69.4 ± 0.8 & 71.0 ± 1.3 & 64.8 ± 1.2 & 63.6 ± 1.6 & 81.7 ± 0.8 & 71.5 ± 0.8 & 79.4 ± 0.6 \\
			& $\mathrm{\mymodel_{cos}}$ & 69.9 ± 0.7 & 70.8 ± 1.4 & 65.2 ± 1.8 & 62.7 ± 1.3 & 82.0 ± 0.7 & 71.9 ± 0.9 & 78.7 ± 0.8 \\
			& $\mathrm{\mymodel_{at}}$ & 69.3 ± 0.8 & 70.9 ± 1.3 & 65.0 ± 1.3 & 65.0 ± 1.7 & 81.9 ± 0.9 & 71.3 ± 0.7 & 78.8 ± 0.6 \\
			& \mymodel & \cellcolor{lightgrey}\textbf{71.8 ±   0.9} & 71.5 ± 0.8 & \cellcolor{lightgrey}\textbf{70.6 ± 1.4} & \cellcolor{lightgrey}\textbf{66.8 ± 1.1} & \cellcolor{lightgrey}\textbf{83.5 ± 0.6} & 73.6 ± 0.6 & 79.8 ± 0.8 \\
			& \mymodelvar & 71.1 ± 0.3 & \cellcolor{lightgrey}\textbf{72.2 ± 0.5} & 70.3 ± 0.9 & \cellcolor{lightgrey}\textbf{66.8 ± 1.4} & \cellcolor{lightgrey}\textbf{83.5 ± 0.6} & \cellcolor{lightgrey}\textbf{73.9 ± 0.7} & \cellcolor{lightgrey}\textbf{79.9 ± 0.5} \\
			\hline
		\end{tabular}
\end{table*}

\subsubsection{Iterative Structure Learning for Acceleration}
A straightforward way is to consider once structure inference and once GNN's message passing for prediction in each feed-forward computation. To enable structure learning and GNN learning mutually reinforce each other~\cite{idgl}, we consider multiple iterative updates of graph structures and node representations before once back-propagation. More specifically, in each epoch, we repeatedly update node representations $\mathbf{Z}^{t} $ (where the superscript $t$ denotes the $t$-th iteration) and latent graph $\hat{\mathbf{A}}^{t} $ until a given maximum budget is achieved. To accelerate the training,
we aggregate the losses $\mathcal L^{t}$ in each iteration step for parameter updating.
As different graphs have different feature space, we utilize the first layer of GNN as an encoder at the very beginning and then feed the encoded representations to structure learner.
The training algorithm for structure learner $g_\theta$ on source graphs is described in Alg.~\ref{alg:meta-learning} (in the appendix) where we train structure learner for multiple episodes and in each episode, we train $g_\theta$ on each source graph for several epochs.
In testing, the well-trained $g_\theta$ is fixed and we train a GNN $h_w$ on the target graph with latent structures inferred by $g_\theta$, as described in Alg.~\ref{alg:sup-learning}. 

\section{Related Works}

\paragraph{Graph Neural Networks} 
Graph neural networks (GNNs)~\cite{gcn,gat,graphsage,jknet-icml18,mixhop-icml19,geng2023pyramid} have achieved impressive performances in modeling graph-structured data. Nonetheless, there is increasing evidence suggesting GNNs' deficiency for graph structures that are inconsistent with the principle of message passing.
One typical situation lies in non-homophilous graphs~\cite{zhu2020beyond}, where adjacent nodes tend to have dissimilar features/labels. Recent studies devise adaptive feature propagation/aggregation to tackle the heterophily~\cite{spectral_filter,zhu2020graph,zhu2020beyond,gpr}. Another situation stems from graphs with noisy or spurious links, for which several works propose to purify the observed structures for more robust node representations~\cite{sparsification,robust-topo}. Our work is related to these works by searching adaptive graph structures that is suitable for GNN's message passing. Yet, the key difference is that our method targets learning a new graph out of the scope of input one, while the above works focus on message passing within the input graph.

\paragraph{Graph Structure learning} To effectively address the limitations of GNNs' feature propagation within observed structures, many recent works attempt to jointly learn graph structures and the GNN model. For instance, \cite{lds} models each edge as a Bernoulli random variable and optimizes graph structures along with the GCN. To exploit enough information from observed structure for structure learning, \cite{rbf} proposes a metric learning approach based on RBF kernel to compute edge probability with node representations, while \cite{attention} adopts attention mechanism to achieve the similar goal. Furthermore,
\cite{idgl} considers an iterative method that enables mutual reinforcement between learning graph structures and node embeddings. Also,
\cite{bgcnn} presents a probabilistic
framework that views the input graph as a random sample from a collection modeled by a parametric random graph model.
\cite{lao2022variational, variational_inference_2} harnesses variational inference to estimate a posterior of graph structures and GNN parameters. While learning graph structures often requires $O(N^2)$ complexity, a recent work \cite{wunodeformer} proposes an efficient Transformer that achieves latent structure learning in each layer with $O(N)$ complexity. However, though these methods have shown promising results, they assume training nodes and testing nodes are from the same graph and consider only one graph.
By contrast, we consider graph structure learning under the cross-graph setting and propose a general framework to learn a shared structure learner which can generalize to target graphs without any re-training.

\paragraph{Out-of-Distribution Generalization on Graphs.} Due to the demand for handling testing data in the wild, improving the capability of the neural networks for performing satisfactorily on out-of-distribution data has received increasing attention~\cite{wu2022handling,ma2021subgroup,ligraphde,wu2023energybased,wu2021towards2,yang2023pmlp}. Recent studies, e.g., \cite{zhu2021shift,wu2022handling,sui2022causal} explore effective treatments for tackling general distribution shifts on graphs, and there are also works focusing on particular categories of distribution shifts like size generalization~\cite{bevilacqua2021size}, molecular scaffold generalization~\cite{yang2022learning}, feature/attribute shifts~\cite{wu2021towards,ood2023bi}, topological shifts~\cite{yang2022geometric}, etc. To the best of our knowledge, there is no prior works considering OOD generalization in the context of graph structure learning. In our case, the target graph, where the structure learner is expected to yield adaptive structures, can have disparate distributions than the source graphs. The distribution shifts could potentially stem from feature/label space, graph sizes or domains (e.g., from social networks to citation networks). As the first attempt along this path, our work can fill the research gap and enable the graph structure learning model to deal with new unseen graphs in an open world.

\section{Experiments}

We apply \mymodel to real-world datasets for node classification to test the efficacy of proposed structure learner for boosting performance of GNN learning on target graphs with distribution shifts from source graphs. We specify the backbone GNN network for \mymodel as a two-layer GCN~\cite{gcn}. We focus on the following research questions: 

\noindent\bb\;\textbf{1)} How does \mymodel perform compared with directly training GNN models on input structure of target graphs? 

\noindent\bb\;\textbf{2)} How does \mymodel perform compared to state-of-the-art structure learning models that are directly trained on target datasets in terms of both accuracy and training time? 

\noindent\bb\;\textbf{3)} Are the proposed components of \mymodel effective and necessary for the achieved performance? 

\noindent\bb\;\textbf{4)} What is the impact of hyper-parameter on performance and what is the impact of attack on observed edges? 

\noindent\bb\;\textbf{5)} What is the property of inferred latent graphs and what generalizable pattern does the structure learner capture?


\subsection{Experimental Protocols}

\textbf{Datasets.} Our experiments are conducted on several public graph datasets. First we consider three commonly used citation networks \textsc{Cora}, \textsc{CiteSeer} and \textsc{PubMed}. We use the same splits as in \cite{planetoid}. These three datasets have high homophily ratios (i.e., adjacent nodes tend to have similar labels) \cite{lim2021new}. Apart from this, we also consider four social networks from \textsc{Facebook-100} \cite{traud2012social}, which have low homophily ratios.
Readers may refer to Appendix \ref{appx-implement} for more dataset information like splitting ratios.

\textbf{Competitors.} We mainly compare with GCN \cite{gcn}, the GNN counterpart trained on input structure, for testing the efficacy of produced latent graphs by \mymodel. As further investigation, we also compare with other advanced GNN models: GraphSAGE \cite{graphsage}, GAT \cite{gat}, APPNP \cite{appnp}, $\mbox{H}_2$GCN \cite{zhu2020beyond} and GPRGNN \cite{gpr}. Here APPNP, $\mbox{H}_2$GCN and GPRGNN are all strong GNN models equipped with adaptive feature propagation and high-order aggregation. For these pure GNN models, the training and testing are considered on (the same) target graphs. 
Furthermore, we compete \mymodel with state-of-the-art graph structure learning models, IDS~\cite{lds}, IDGL~\cite{idgl} and VGCN~\cite{variational_inference_2}. Since these models are all designed for training on one dataset from scratch, we directly train them on the target graph and they in principle could yield better performance than \mymodel.

We also consider variants of \mymodel as baselines. We replace the similarity function $s$ with attention-based structure learner, denoted as \mymodelgat, which follows the same training scheme as \mymodel. Besides, we consider some non-parametric similarity functions like dot-product, KNN and cosine distance (denoted as $\mathrm{\mymodel_{dp}}$, $\mathrm{\mymodel_{knn}}$ and $\mathrm{\mymodel_{cos}}$, respectively). For these models, we only need to train the GNN network on target graphs with the non-parametric structure learners yielding latent structures. In addition, we introduce a variant \mymodelvar that shares the same architecture as \mymodel and is directly trained on target graphs. Also, \mymodelvar in principle could produce superior results than \mymodel. We report the test accuracy given by the model that produces the highest validation accuracy within 500 training epochs.

\subsection{In-domain Generalization}

\begin{table*}[tb!]
	\caption{Test accuracy (\%) on target graphs for cross-domain generalizations. For each social network (resp. citation network) as target dataset, we consider citation networks (resp. social networks) as source graphs.}
	\label{tab-res2}
	\begin{tabular}{| c| lccccccc |}
		\hline
		\textbf{Type} & \textbf{Method }& \textbf{Cornell5} & \textbf{Johns.55} & \textbf{Amherst41} & \textbf{Reed98} & \textbf{Cora} & \textbf{CiteSeer} & \textbf{PubMed} \\
		\hline
		\multirow{7}{*}{\specialcell[t]{\textbf{Pure}\\\textbf{GNN}} }& GCN & 68.6 ± 0.5 & 70.8 ± 1.0 & 65.8 ± 1.6 & 60.8 ± 1.6 & 81.6 ± 0.4 & 71.6 ± 0.3 & 78.8 ± 0.6 \\
		& SAGE & 68.7 ± 0.8 & 67.5 ± 0.9 & 66.3 ± 1.8 & 63.9 ± 1.9 & 81.4 ± 0.6 & 71.6 ± 0.5 & 78.6 ± 0.7 \\
		& GAT & 69.6 ± 1.2 & 69.4 ± 0.7 & 68.7 ± 2.1 & 64.5 ± 2.5 & 83.0 ± 0.7 & 72.1 ± 1.1 & 79.0 ± 0.4 \\
		& GPR & 68.8 ± 0.7 & 69.6 ± 1.3 & 66.2 ± 1.5 & 62.7 ± 2.0 & 83.1 ± 0.7 & 72.4 ± 0.8 & 79.6 ± 0.5 \\
		& APPNP & 68.5 ± 0.8 & 69.1 ± 1.4 & 65.9 ± 1.3 & 62.3 ± 1.5 & 82.7 ± 0.5 & 71.9 ± 0.5 & 79.2 ± 0.3 \\
		& H$_2$GCN & 71.4 ± 0.5 & 68.3 ± 1.0 & 66.5 ± 2.2 & 65.4 ± 1.3 & 82.5 ± 0.8 & 71.4 ± 0.7 & 79.4 ± 0.4 \\
		& CPGNN & 71.1 ± 0.5 & 68.7 ± 1.3 & 66.7 ± 0.8 & 63.6 ± 1.8 & 80.8 ± 0.4 & 71.6 ± 0.4 & 78.5 ± 0.7 \\
		\hline
		\multirow{6}{*}{\specialcell[t]{\textbf{Graph}\\\textbf{Structure}\\\textbf{Learning}}} & $\mathrm{\mymodel_{dp}}$ & 71.5 ± 0.7 & 71.3 ± 1.2 & 68.5 ± 1.6 & 63.2 ± 1.2 & 83.1 ± 0.8 & 71.7 ± 1.0 & 77.3 ± 0.8 \\
		& $\mathrm{\mymodel_{knn}}$ & 69.4 ± 0.8 & 71.0 ± 1.3 & 64.8 ± 1.2 & 63.6 ± 1.6 & 81.7 ± 0.8 & 71.5 ± 0.8 & 79.4 ± 0.6 \\
		& $\mathrm{\mymodel_{cos}}$ & 69.9 ± 0.7 & 70.8 ± 1.4 & 65.2 ± 1.8 & 62.7 ± 1.3 & 82.0 ± 0.7 & 71.9 ± 0.9 & 78.7 ± 0.8 \\
		& $\mathrm{\mymodel_{at}}$ & 69.9 ± 1.0 & 70.4 ± 1.5 & 64.4 ± 1.2 & 65.0 ± 1.7 & 82.5 ± 0.9 & 71.8 ± 0.8 & 78.5 ± 0.7 \\
		& \mymodel & \cellcolor{lightgrey}\textbf{72.0 ± 1.0} & 71.8 ± 0.7 & 69.8 ± 1.3 & \cellcolor{lightgrey}\textbf{67.3 ± 1.2} & 83.2 ± 0.4 & 73.8 ± 0.9 & 79.6 ± 0.7 \\
		& \mymodelvar & 71.1 ± 0.3 & \cellcolor{lightgrey}\textbf{72.2 ± 0.5} & \cellcolor{lightgrey}\textbf{70.3 ± 0.9} & 66.8 ± 1.4 & \cellcolor{lightgrey}\textbf{83.5 ± 0.6} & \cellcolor{lightgrey}\textbf{73.9 ± 0.7} & \cellcolor{lightgrey}\textbf{79.9 ± 0.5} \\
		\hline
	\end{tabular}	
\end{table*}

\begin{figure*}[t]
\subfigure[Cora]{
\includegraphics[width=0.19\textwidth]{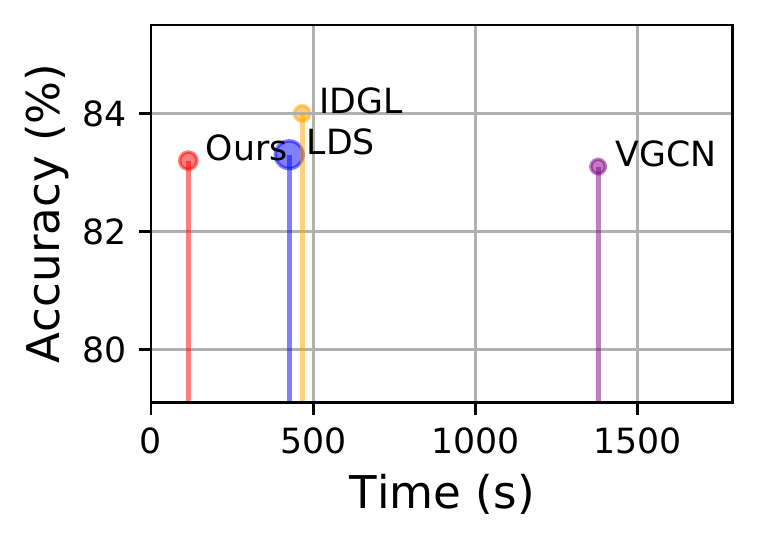}}
\subfigure[CiteSeer]{
\includegraphics[width=0.19\textwidth]{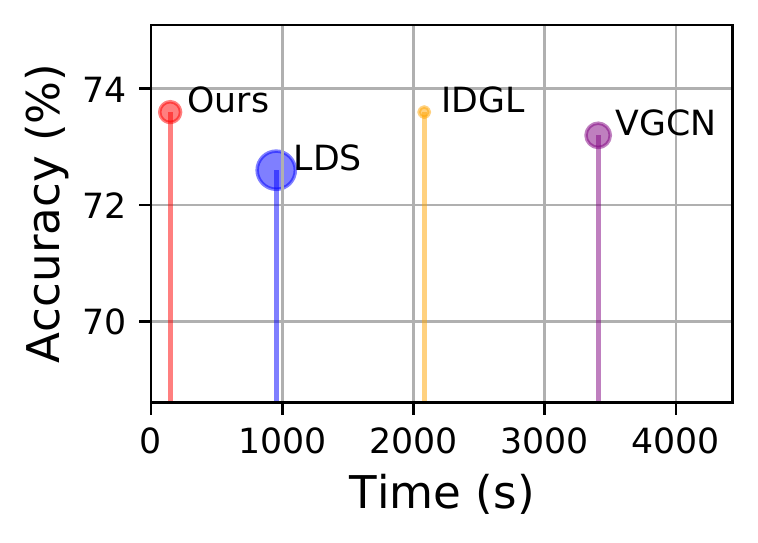}}
\subfigure[Amherst41]{
\includegraphics[width=0.19\textwidth]{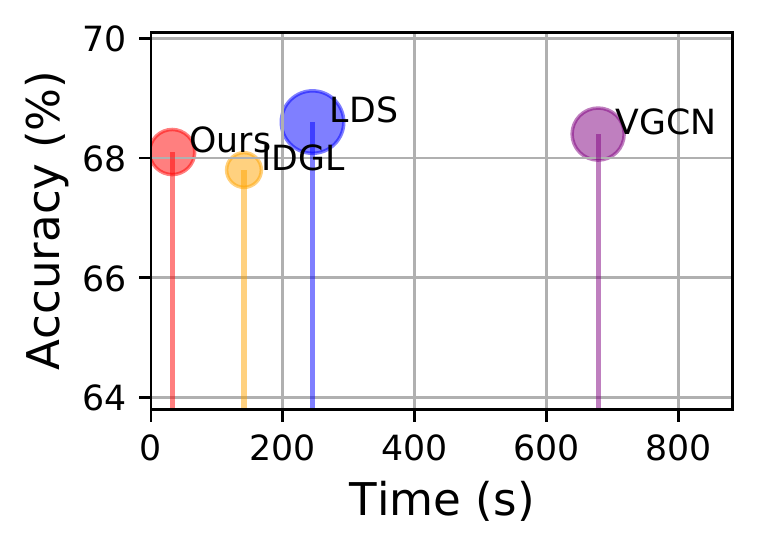}}
\subfigure[Johns Hopkins55]{
\includegraphics[width=0.187\textwidth]{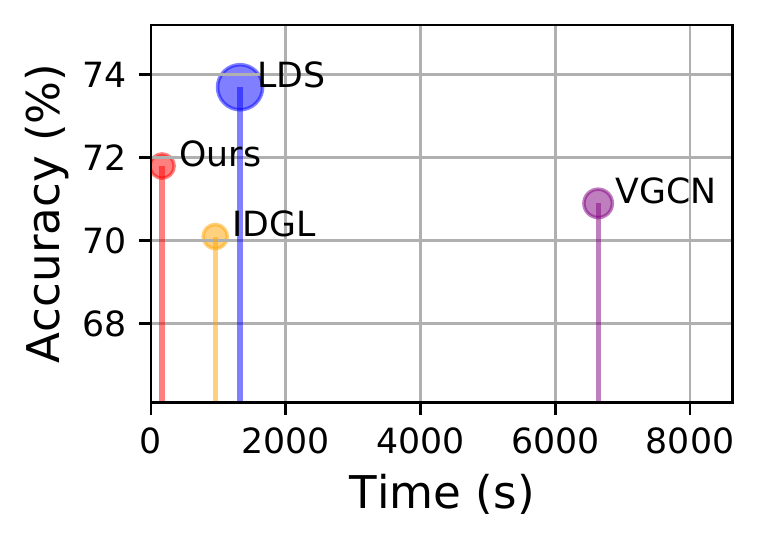}}
\subfigure[Reed98]{
\includegraphics[width=0.19\textwidth]{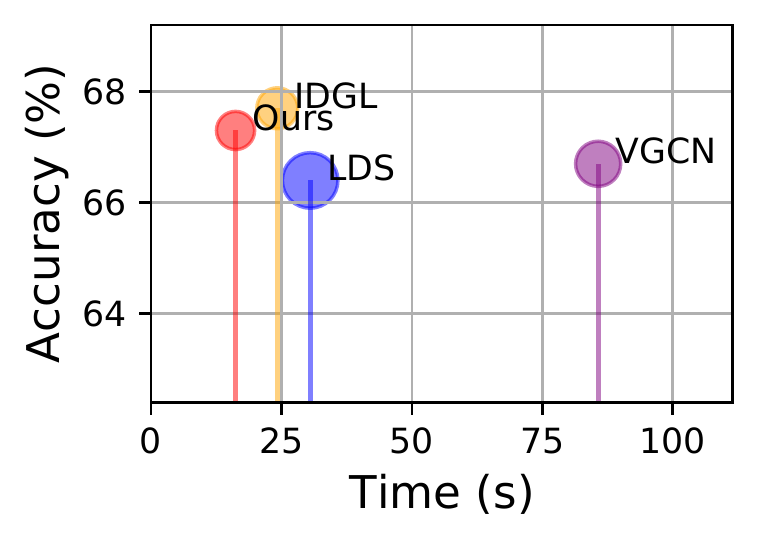}}
\caption{Comparison of test accuracy and training time with SOTA structure learning models (LDS~\cite{lds}, IDGL~\cite{idgl} and VGCN~\cite{variational_inference_2}). The radius of circle is proportional to standard deviation. The experiments are run on one Tesla V4 with 16 GPU memory. We adopt the same setting as Table~\ref{tab-res2} and report the results on target datasets. For \textsc{Cornell5} and \textsc{PubMed}, the competitor models suffer out-of-memory.}
\label{fig-res}
\end{figure*}

We first consider transferring within social networks or citation networks. The results are reported in Table~\ref{tab-res1} where for each social network (resp. citation network) as the target, we use the other social networks (resp. citation networks) as the source datasets.  
\mymodel performs consistently better than GCN, i.e., the counterpart using observed graph for message passing, which proves that \mymodel can capture generalizable patterns for desirable message-passing structure for unseen datasets that can indeed boost the GCN backbone's performance on downstream tasks. In particular, the improvement over GCN is over 5\% on \textsc{Cornell5} and \textsc{Reed98}, two datasets with low homophily ratios (as shown in Table~\ref{tab:dataset_statistic}). The reason is that for non-homophilous graphs where the message passing may propagate inconsistent signals (as mentioned in Section~\ref{sec-intro}), the GNN learning could better benefits from structure learning than homophilous graphs. Furthermore, compared to other strong GNN models, \mymodel still achieves slight improvement than the best competitors though the backbone GCN network is less expressive. One could expect further performance gain by \mymodel if we specify the GNN backbone as other advanced architectures.

In contrast with non-parametric structure learning models and \mymodelgat, \mymodel outperforms them by a large margin throughout all cases, which verifies the superiority of our design of multi-head weighted similarity function that can accommodate multi-faceted diverse structural information. Compared with \mymodelvar, \mymodel performs on par with and even exceeds it on \textsc{Cornell5} and \textsc{Amherst41}. The possible reasons are two-fold. First, there exist sufficient shared patterns among citation networks (resp. social networks), which paves the way for successful generalization of \mymodel. Second, \mymodelvar could sometimes overfit specific datasets, since the amount of free parameters are regularly orders-of-magnitude more than the number of labeled nodes in the dataset. The results also imply that our transfer learning approach can help to mitigate over-fitting on one dataset. 
Moreover, \mymodel can generalize structure learner to unseen graphs that is nearly three times larger than training graphs, i.e., \textsc{Cornell5}. 

\subsection{Cross-domain Generalization}
We next consider a more difficult task, transferring between social networks and citation networks. The difficulty stems from two aspects: 1) social networks and citations graphs are from distinct categories thus have larger underlying data-generating distribution gaps; 2) they have varied homophily ratios, which indicates that the observed edges play different roles in original graphs. 
In Table~\ref{tab-res2} we report the results.
Despite the task difficulty, \mymodel manages to achieve superior results than GCN and also outperforms other non-parametric graph structure learning methods throughout all cases. This suggests \mymodel's ability for handling target graphs with distinct properties. 

In Fig.~\ref{fig-res} we further compare \mymodel with three state-of-the-art graph structure learning models that are directly trained on target graphs. Here we follow the setting in Table~\ref{tab-res2}. The results show that even trained on source graphs that are different from the target one, \mymodel still performs on par with the competitors that are trained and tested on (the same) target graphs. Notably, \mymodel significantly reduces training time. 
For instance, in \textsc{John Hopkins55}, \mymodel is 6x, 9x and 40x faster than IDGL, LDS and VGCN, respectively. This shows one clear advantage of \mymodel in terms of training efficiency and also verifies that our model indeed helps to reduce the significant cost of training time for structure learning on target graphs.

\subsection{Ablation Studies}

We conduct ablation studies to test the effectiveness of iterative learning scheme and regularization on graphs.

\begin{figure}[t]
	\centering
	\subfigure[]{
		\begin{minipage}[t]{0.51\linewidth}
			\centering
			\includegraphics[width=0.9\textwidth,angle=0]{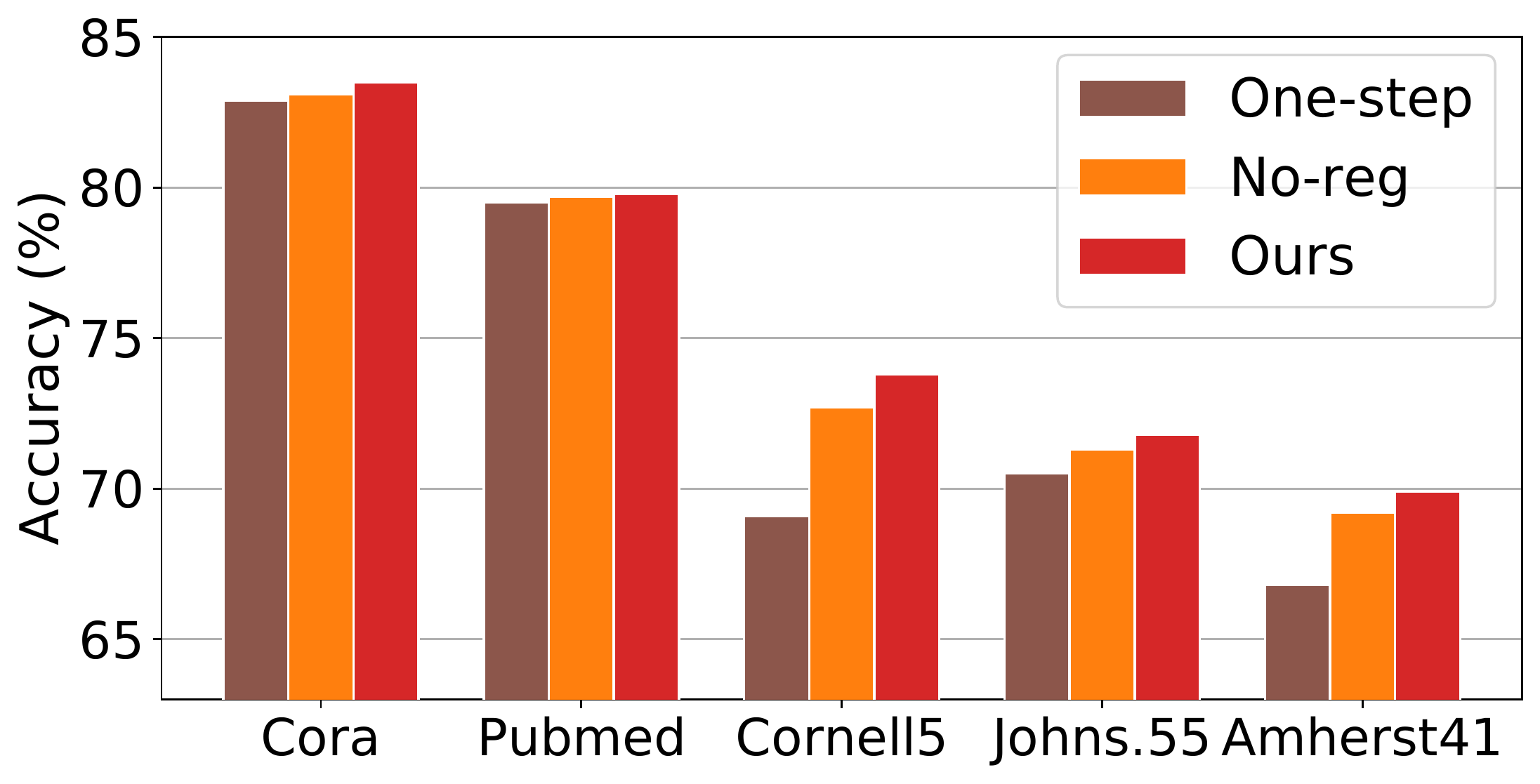}
			\label{fig:ablation}
		\end{minipage}	
}
	\subfigure[]{
		\begin{minipage}[t]{0.43\linewidth}
			\centering
\includegraphics[width=0.9\textwidth,angle=0]{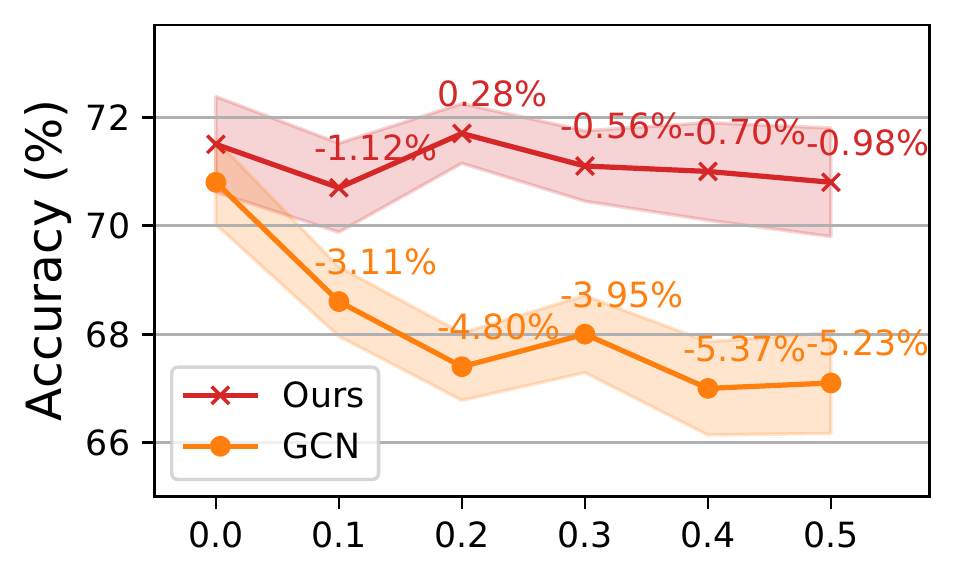}
\label{fig:edge_delete}
		\end{minipage}
	}
	\caption{(a) Ablation study for \mymodel. (b) Performance comparison of \mymodel and GCN w.r.t. randomly removing certain ratios of edges in Johns Hopkins55.}
\end{figure}

\textbf{Effect of Iterative Learning.}
We replace the iterative learning process as a one-step prediction (i.e., once structure estimation and updating node representations in once feed-forward computation) and compare its test accuracy with \mymodel. The results are shown in Fig. ~\ref{fig:ablation} where we follow the setting of Table~\ref{tab-res1}. 
The non-iterative version exhibits a considerable drop in accuracy (as large as 5.4\% and 8.8\% when tested on target graphs \textsc{Cornell5} and \textsc{Amherst41}, respectively).
Therefore, the iterative updates indeed help to learn better graph structures and node embeddings, contributing to higher accuracy for downstream prediction.

\textbf{Effect of Regularization on structures.}
We remove the regularization on structures (i.e., setting $\alpha = \rho = 0$) and compare with \mymodel. As shown in Fig.~\ref{fig:ablation}, there is more or loss performance degradation. In fact, the regularization loss derived from the prior distribution for latent structures could help to provide some guidance for structure learning, especially when labeled information is limited.

\begin{figure}[tb!]
	\centering
	\subfigure[]{
		\begin{minipage}[t]{0.5\linewidth}
			\centering
			\includegraphics[width=0.9\textwidth,angle=0]{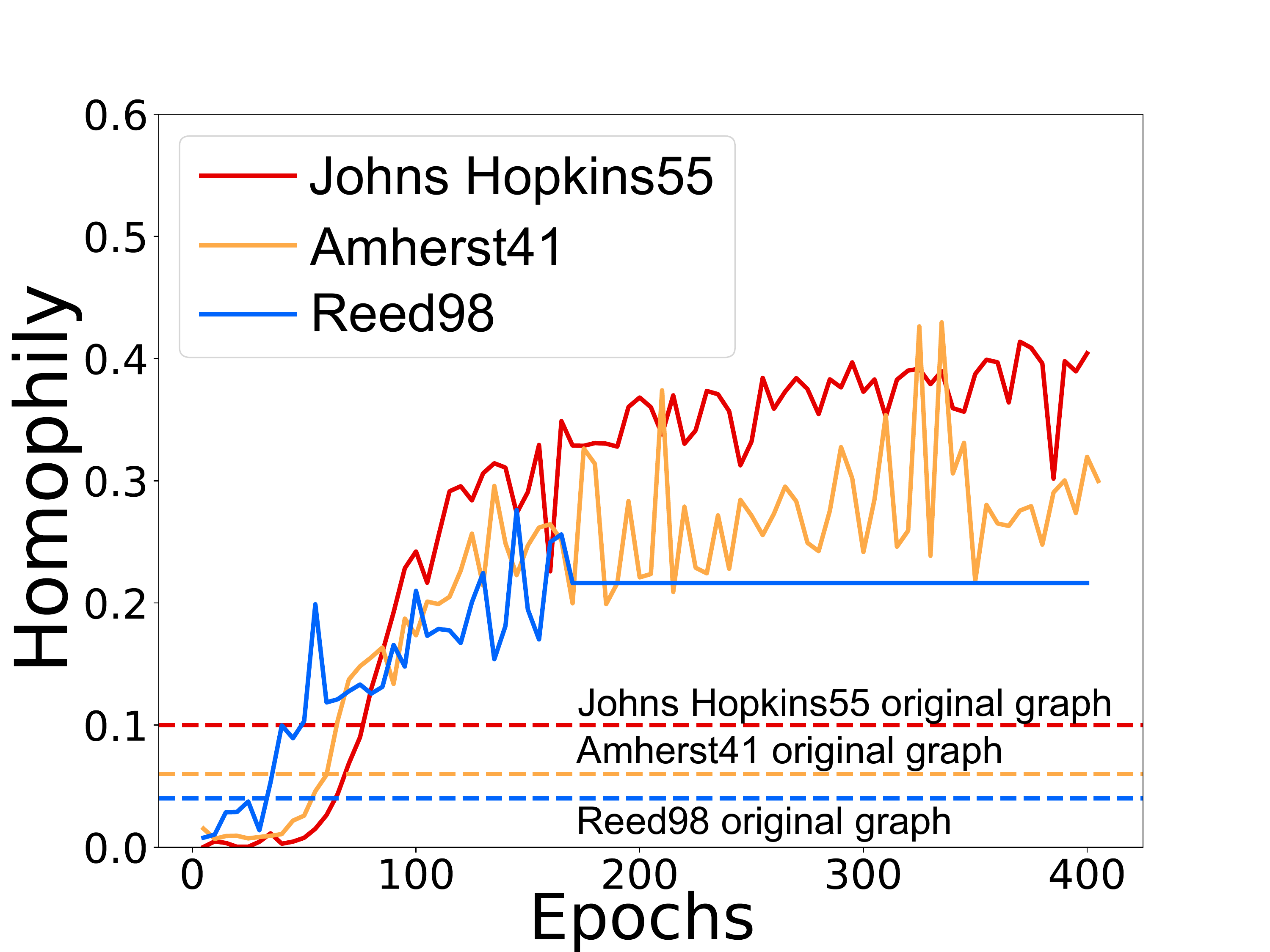}
			\label{fig:homo}
		\end{minipage}%
	}%
	\subfigure[]{
		\begin{minipage}[t]{0.5\linewidth}
			\centering
			\includegraphics[width=0.9\textwidth,angle=0]{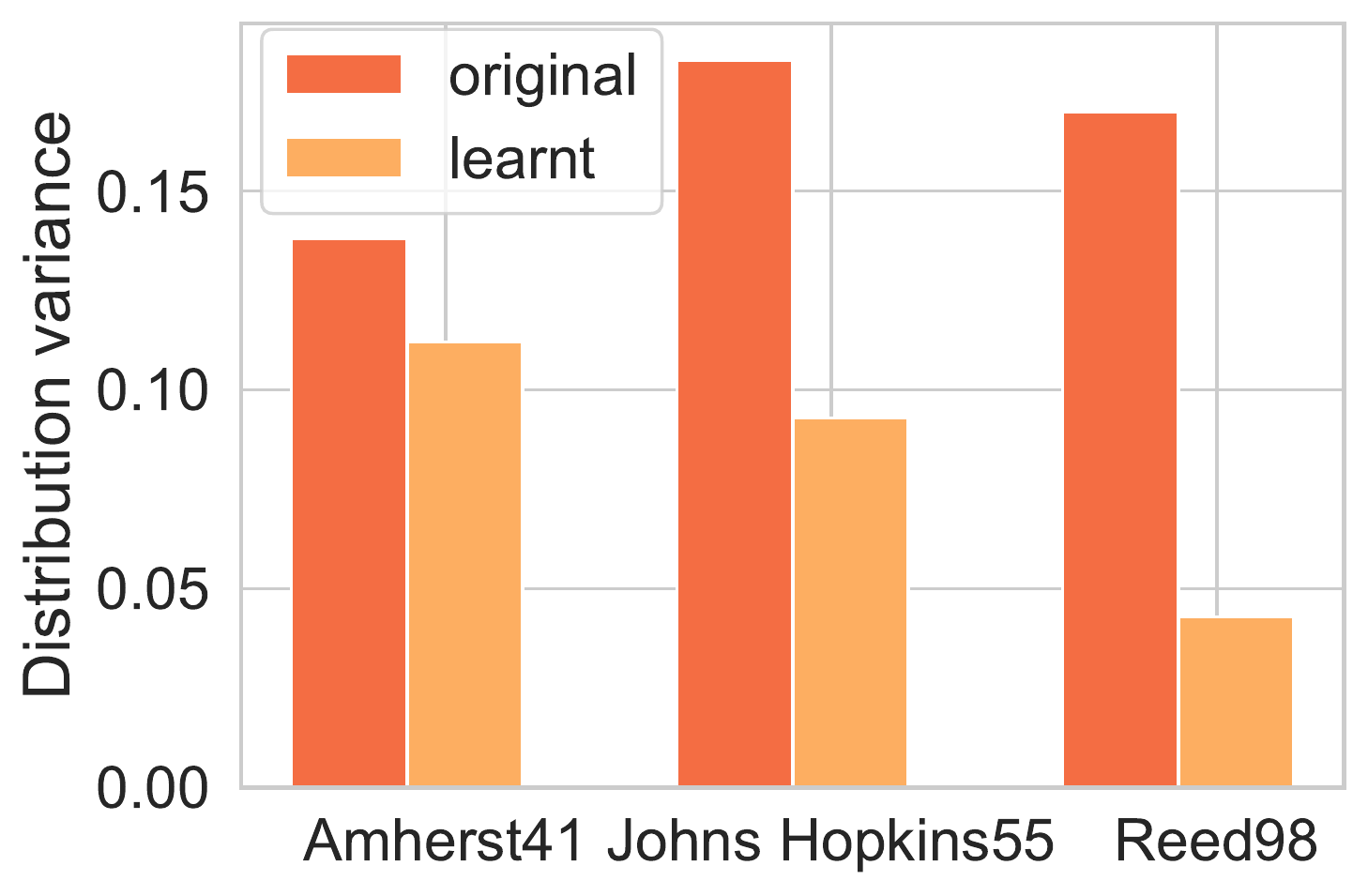}
			\label{fig:variance}
		\end{minipage}%
	}%
	\caption{(a) The curves of homophily ratios for latent structures during the learning process. (b)  The variance of neighborhood distribution of nodes with the same label in original graphs and learnt structure. }
	\label{fig}
\end{figure}

\subsection{Hyper-parameter Sensitivity}

In Fig.~\ref{fig-para} (in the appendix), we study the variation of model's performance w.r.t. $\lambda$ (the weight on input graphs) and $P$ (the number of pivots) on target datasets \textsc{Cora} and \textsc{CiteSeer}. 
Overall, the model is not sensitive to $\lambda$'s.
For \textsc{Cora}, larger $\lambda$ contributes to higher accuracy, while for \textsc{CiteSeer}, smaller $\lambda$ yields better performance. The possible reason is that the initial graph of \textsc{Cora} is more suitable for message passing (due to higher homophily ratio).
For the impact of pivot number, as shown in Fig.~\ref{fig-para}(b), a moderate value of $P$ could provide decent downstream performance. 

\subsection{Robustness Analysis}

In addition, we find that \mymodel is more immune to edge deletion attack than GCN. We randomly remove 10-50\% edges of target graphs respectively, and then apply \mymodel and GCN. 
We present the results in Johns Hopkins55 in Fig.~\ref{fig:edge_delete} and leave more results in Appendix \ref{appx-moreres}. When the drop ratio increases, the performance gap between two models becomes more significant.
This is due to our structure learner's ability for learning new graph structures from node embeddings, making it less reliant on initial graph structures and more robust to attack on input edges. 

\subsection{Case Study}
We further probe into why our approach is effective for node classification by dissecting the learnt graph structures. 
Specifically, we measure the homophily ratios of learnt structures and their variance of neighborhood distributions of nodes with same labels.
As nodes receive messages from neighbors in message passing, the more similar the neighborhood patterns of nodes within one class are, the easier it is for GNNs to correctly classify them \cite{ma2022is}.
We use homophily metric proposed in \cite{lim2021new}  to measure homophily ratios.
For calculation of variance of neighborhood distribution, we first calculate variance for each class, and then take weighted sum to get the final variance, where the weight is proportional to the number of nodes within corresponding class.

\textbf{Homophily Ratio.} We choose \textsc{Amherst41}, \textsc{Johns Hopkins55} and \textsc{Reed98} as target graphs, and record the homophily ratios of inferred latent structures every five epochs during training. As shown in Fig. \ref{fig:homo}. the homophily ratios of inferred latent graphs exhibit a clear increase as the training epochs become more and the final ratio is considerably larger than that of input graph. The results indicate that the trained structure learner incline to output more homophilous latent structures that are reckoned to be more suitable for message passing. 

\textbf{Neighborhood Distribution Variance.} As shown in Fig. \ref{fig:variance}, the variance of neighborhood distribution of nodes with the same label is  significantly smaller in our learnt structure, making it easier to classify nodes through message passing.
The results also imply that  high homophily ratio and similar intra-class neighborhood patterns could be two of the underlying transferrable patterns of optimal message-passing structure, identified by \mymodel.

\section{Conclusion}
This paper proposes \textit{Graph Structure Learning Under Cross-Graph Distribution Shift}, a new problem that requires structure learner to transfer to new target graphs without re-training and handles distribution shift.
We develop a transfer learning framework that guides the structure learner to discover shared knowledge across source datasets with respect to optimal message-passing structure for boosting downstream performance. We also carefully design the model components and training approach in terms of expressiveness, scalability and stability.
We devise experiments with various difficulties and  demonstrate the efficacy and robustness of our approach.
Although our framework is pretty general, we believe their are other potential methods that can lead to equally competitive results, which we leave as future work.

\begin{acks}
The work was supported in part by National Key Research and Development Program of China (2020AAA0107600), NSFC (62222607), Science and Technology Commission of Shanghai Municipality (22511105100), and Shanghai Municipal Science and Technology Major Project (2021SHZDZX0102).
\end{acks}
	
\clearpage

\bibliographystyle{ACM-Reference-Format}
\balance
\bibliography{ref}


\appendix

\clearpage
\nobalance

\begin{algorithm}[t]
    \caption{Message Passing over Latent Graphs $[\mathbf Z, \hat{\mathbf Y}] = \mbox{GNN}(\mathbf A, \mathbf X, \mathbf \Gamma; w)$}
    \label{alg:GNN}
    \KwIn{node features $\mathbf X$, input adjacency $\mathbf A$, latent graph node-pivot similarity matrix $\mathbf \Gamma = \{\alpha_{up}\}_{N\times P}$.}
	$\mathbf Z^{(0)} \leftarrow \mathbf X$\;
	\For {$l = 0,1,\cdots,L-1$}{
	$\mathbf Z^{(l+1)} \leftarrow \mathbf D^{-\frac{1}{2}} \mathbf A \mathbf D^{-\frac{1}{2}} \mathbf Z^{(l)} \mathbf W^{(l)}$\;
	$\mathbf C^{(l+\frac{1}{2})} = \mathrm{RowNorm}  (\mathbf \Gamma^\top) \mathbf Z^{(l)}$\;
	$\mathbf C^{(l+1)} = \mathrm{RowNorm} (\mathbf \Gamma) \mathbf C^{(l+\frac{1}{2})}$ \;
	$\mathbf Z^{(l+1)} \leftarrow \sigma \left((\lambda \mathbf Z^{(l+1)}+ (1-\lambda)\mathbf C^{(l+1)} \mathbf W^{(l)} \right) $\;
	}
	\KwOut{node representations $\mathbf Z = \mathbf Z^{(L-1)}$, node-level prediction $\hat{\mathbf Y} = \mathbf Z^{(L)}$.}
\end{algorithm}

\begin{algorithm}[t]
	\caption{Training Graph Structure Learner}
	\label{alg:meta-learning}
	\KwIn{observed source graphs $\{\mathcal G_m\}_{m=1}^M = \{(\mathbf A_m, \mathbf X_m, \mathbf Y_m)\}_{m=1}^M$, maximum training episode $M$, maximum iteration $T$, a shared graph structure learner $g_\theta$, GNN networks $ \{ h_{w_m} \}_{m=1}^M$ for each source graph.}
	Initialize $\theta$ and $\{w_m\}_{m=1}^M$\;
	\For {episode: $1,2,\dots,E$}{
		\For {each source graph $\mathcal G_m$}{
			\For {$e=1:T$}{
				$\Delta_\theta \leftarrow 0$, $\Delta_w \leftarrow 0$, $ t \leftarrow 0 $\;
				$\mathbf Z^0 = \mbox{MLP}(\mathbf X; w_m)$ or $\mathbf Z^0 = \mbox{GCN}(\mathbf A, \mathbf X; w_m)$\;
				\While{not converged}{
					$ t \leftarrow t+1 $\;
					Compute $\mathbf \Gamma^t = \{\alpha_{up}\}$ using \eqref{eqn-sim} with $\mathbf Z^{t-1}$\;
					Sample $K$ times over $\mathbf \Gamma^t$ to obtain $\{\hat{\mathbf E}_k^t\}_{k=1}^K$\;
					Compute $\{\pi_\theta(\hat{\mathbf E}_k^t)\}_{k=1}^K$ using \eqref{eqn-pi}\;
					$[\mathbf Z^t, \hat{\mathbf Y}^t] = \mbox{GNN}(\mathbf A, \mathbf X, \mathbf \Gamma^t; w_m)$\;
					Compute $\nabla_{\theta} \mathcal L_s, \nabla_{\theta} \mathcal L_r, \nabla_{\theta} \mathcal L_e$ using \eqref{eqn-loss-sup-grad}, \eqref{eqn-grad-reg}, \eqref{eqn-loss-ent}, respectively \;
					$\Delta_\theta^{(t)} \leftarrow \nabla_{\theta} \mathcal L_s + \nabla_{\theta} \mathcal L_r + \nabla_{\theta} \mathcal L_e$\;
					$\Delta_w^{(t)} \leftarrow \nabla_{w_m} \mathcal L_s$\;
				}
				$\Delta_\theta \leftarrow \Delta_\theta^{(0)} + \sum_{i=2}^t \Delta_\theta^{(i)} / (t-1)  $\;
				$\Delta_w \leftarrow \Delta_w^{(0)} +  \sum_{i=2}^t\Delta_w^{(i)} / (t-1)  $\;
				Use gradient $\Delta_\theta$ to update $\theta$\;
				Use gradient $\Delta_w$ to update $w_m$\;
			}
		}
	}
	\KwOut{trained graph structure learner $g_{\theta^*}$.}
\end{algorithm}

\begin{algorithm}[t]
    \caption{Supervised Learning for target GNN}
    \label{alg:sup-learning}
    \KwIn{observed target graphs $\mathcal G = (\mathbf A, \mathbf X, \mathbf Y)$, GNN network $h_w$ for the target graph, maximum iteration $T$.}
	Initialize $w$\;
	\For {$e=1:T$}{
	$t \leftarrow 0$, $\Delta_w \leftarrow 0$, $\mathbf Z^{(0)} \leftarrow \mathbf X_m$, $\mathbf A \leftarrow \mathbf A_m$\;
	$\mathbf Z^0 = \mbox{MLP}(\mathbf X; w)$ or $\mathbf Z^0 = \mbox{GCN}(\mathbf A, \mathbf X; w)$\;
	\While{not converged}{
				$ t \leftarrow t+1 $\;
			    Compute $\mathbf \Gamma^t = \{\alpha_{up}\}$ using \eqref{eqn-sim} with $\mathbf Z^{t-1}$\;
				Sample $K$ times over $\mathbf \Gamma^t$ to obtain $\{\hat{\mathbf E}^t_k\}_{k=1}^K$\;
				Compute $\{\pi_\theta(\hat{\mathbf E}^t_k)\}_{k=1}^K$ using \eqref{eqn-pi}\;
				$[\mathbf Z, \hat{\mathbf Y}] = \mbox{GNN}(\mathbf A, \mathbf X, \mathbf \Gamma^t; w)$\;
				Compute $\nabla_{\theta} \mathcal L_s$ using \eqref{eqn-loss-sup-grad} \;
				$\Delta_w^{(t)} \leftarrow \nabla_{w_m} \mathcal L_s$\;
			}
		$\Delta_w \leftarrow \Delta_w^{(0)} +  \sum_{i=2}^t\Delta_w^{(i)} / (t-1)  $\;
		Use gradient $\Delta_w$ to update $w$\;
		}
	\KwOut{trained GNN network $h_{w^*}$ for the target graph.}
\end{algorithm}

\section{Derivations for NWGM}\label{appx-deri}

First, when taking the gradient, we have $\nabla_\theta \mathbb E_{q_\theta} [ \log p_{w_m}(\mathbf Y| \mathbf A, \mathbf X, \hat{\mathbf A})]$ $ \approx c \nabla_\theta   \log  \mathbb E_{q_\theta} [ p_{w_m}(\mathbf Y| \mathbf A, \mathbf X, \hat{\mathbf A})]$ with basic applications of the chain rule. We then adopt Normalized Weighted Geometric Mean (NWGM)~\cite{NWGM}:
\begin{equation}
    \begin{split}
        & \nabla_\theta   \log \mathbb E_{q_\theta(\hat{\mathbf A}|\mathbf A, \mathbf X)} [ p_{w_m}(\mathbf Y_{u, c} = 1| \mathbf A, \mathbf X, \hat{\mathbf A})] \\ 
        = & \nabla_\theta \log \sum_{\hat{\mathbf A}} \frac{\exp(s_1(\hat{\mathbf A}))}{\exp(s_1(\hat{\mathbf A})) + \exp(s_2(\hat{\mathbf A}))} q_\theta(\hat{\mathbf A}|\mathbf A, \mathbf X) \\
        = & \nabla_\theta \log \sum_{\hat{\mathbf A}} \mbox{Softmax}(s_1(\hat{\mathbf A})) q_\theta(\hat{\mathbf A}|\mathbf A, \mathbf X)  \\
        \approx &  \nabla_\theta \log \mbox{NWGM}(\mbox{Softmax}(s_1(\hat{\mathbf A}))) \\
        = & \nabla_\theta \log \sum_{\hat{\mathbf A}} \frac{\prod_{\hat{\mathbf A}} [\exp (s_1(\hat{\mathbf A}))]^{q_\theta(\hat{\mathbf A}|\mathbf A, \mathbf X)}}{ \prod_{\hat{\mathbf A}}\exp (s_1(\hat{\mathbf A}))]^{q_\theta(\hat{\mathbf A}|\mathbf A, \mathbf X)} + \prod_{\hat{\mathbf A}}\exp (s_2(\hat{\mathbf A}))]^{q_\theta(\hat{\mathbf A}|\mathbf A, \mathbf X)} } \\
        = & \nabla_\theta \log \sum_{\hat{\mathbf A}} \frac{\exp(\sum_{\hat{\mathbf A}} s_1(\hat{\mathbf A})q_\theta(\hat{\mathbf A}|\mathbf A, \mathbf X))}{ \exp(\sum_{\hat{\mathbf A}} s_1(\hat{\mathbf A})q_\theta(\hat{\mathbf A}|\mathbf A, \mathbf X)) + \exp(\sum_{\hat{\mathbf A}} s_2(\hat{\mathbf A})q_\theta(\hat{\mathbf A}|\mathbf A, \mathbf X))}\\
        = & \nabla_\theta \log \sum_{\hat{\mathbf A}} \frac{\exp(\mathbb E_{q_\theta(\hat{\mathbf A}|\mathbf A, \mathbf X)} [s_1(\hat{\mathbf A})] )}{ \exp(\mathbb E_{q_\theta(\hat{\mathbf A}|\mathbf A, \mathbf X)} [s_1(\hat{\mathbf A})] ) + \exp(\mathbb E_{q_\theta(\hat{\mathbf A}|\mathbf A, \mathbf X)} [s_2(\hat{\mathbf A})] ) } \\
        = & \nabla_{\theta}  \log p_{w_m} (\mathbf Y_{u,c} = 1| \mathbf A, \mathbf X, \hat{\mathbf A} = \mathbb E_{q_\theta(\hat{\mathbf A}|\mathbf A, \mathbf X)} [\hat{\mathbf A}]),
    \end{split}
\end{equation}
where $s_1$ denotes the positive predicted score for class $c$ which is indeed associated with node $u$, and $s_2(\hat{\mathbf A}) = 0$ in our case. We thus conclude the proof for \eqref{eqn-loss-sup-grad}.

\begin{table}[tb!]
	\caption{Statistic information on experimental datasets. The column \textit{Homo.} reports the homophily ratios, measured by the metric proposed in \cite{lim2021new}.}
	\label{tab:dataset_statistic}
	\begin{center}
		\small
		\begin{tabular}{lcccc} 
			\toprule 
			\textbf{Dataset}&
			\textbf{Type}&
			
			\textbf{\# Node }&
			\textbf{\# Edge }&
			\textbf{Homo.}\\
			\midrule 
			Cora & citation & 
			2,708 & 5,429   & 0.77 \\ 
			CiteSeer & citation &
			3,327 & 4,732  &  0.63 \\
			PubMed& citation &
			
			19,717 & 44,338   &  0.66 \\
			Amherst41& social & 
			2,235 & 90,964   & 0.06  \\
			Cornell5& social & 
			18,660 & 790,777   & 0.09  \\
			Johns Hopkins55& social  &
			5,180 & 186,586   & 0.10  \\
			Reed98& social  &
			962 & 18,812   & 0.04 \\
			\bottomrule 
		\end{tabular}
	\end{center}
\end{table}

\section{Datasets and Experimental Details}\label{appx-implement}
The statistical information on datasets is displayed in Table \ref{tab:dataset_statistic}. For the splitting of Cora, CiteSeer and PubMed, we follow \cite{planetoid} to randomly select 20 instances per
class for training, 500/1000 instances for validation/testing in each dataset. In the remaining datasets, 
we employ random train/valid/test splits of 50\%/25\%/25\%. 

The backbone GNN network is specified as a two-layer GCN model. 
We set the similarity function $s$ in \eqref{eqn-sim} as cosine similarity and $\delta$ as a threshold-based truncation. 
Besides, since the dimensions of input node features are different across datasets, we adopt a transformation network that converts input features into a $d$-dimensional node representations before the structure learning module as shown in Alg.~\ref{alg:meta-learning} ($\mathbf Z^0 = \mbox{MLP}(\mathbf X; w_m)$ or $\mathbf Z^0 = \mbox{GCN}(\mathbf A, \mathbf X; w_m)$). We can specify the transformation as a one-layer MLP or a one-layer GCN network (what we adopt). 
Most of the experiments were conducted on an NVIDIA GeForce RTX 2080 Ti with 11GB memory. For experiments involving two larger datasets, PubMed and Cornell5, we utilized an NVIDIA GeForce RTX 3090 with 24 GB memory.

\section{Hyperparameters}\label{appx-hyper}
\label{append:hyper}
We use grid search on validation set to tune the hyperparameters.
The learning rate is searched in $\{0.001, 0.005, 0.01,0.05 \}$; Dropout is searched in $\{0, 0.2, 0.3, 0.5, 0.6 \}$; Hidden channels is searched in $\{16, 32, 64, 96\}$.
Other hyperparameters for specific models are stated below.

For GCN, GraphSAGE and H$^2$GCN, we use 2 layers. For GAT, we search gat head number in $\{2, 4 \}$ and use 2 layers. For APPNP and GPR, we search $\alpha$ in $\{0.1, 0.2, 0.5\}$ and set $K$ to 10. We list the searching space of structure learning methods below.
\begin{itemize}	
	\item
	\mymodel and its variants:
	pivot number $P \in$ \{800, 1000, 1200, 1400\}, embedding size $d \in$ \{16, 32, 64, 96\},
	$\lambda \in$ [0.1, 0.9], 
	$\alpha \in$ \{0, 0.1, 0.15, 0.2, 0.25, 0.3\}, $\rho \in $ \{0, 0.1, 0.15, 0.2, 0.25, 0.3\}, threshold $\in $ \{4e-5, 8.5e-5\}, $H \in $ \{4, 6\}, $T=10$,  $E\in \{1, 2, 3\}$. 

	\item 
	LDS: the sampling time $S=16$, the patience window size $\rho \in $\{10, 20\}, the hidden size $\in$ \{8, 16, 32, 64\}, 
    the inner learning rate $\gamma \in$ \{1e-4, 1e-3, 1e-2, 1e-1\}, 
    and the number of updates used to compute the truncated hypergradient $\tau \in$ \{5, 10, 15\}. 
	\item
	IDGL: $\epsilon=0.01$, hidden size $\in$ \{16, 64, 96,128\}, $\lambda \in$ \{0.5, 0.6, 0.7, 0.8\}, $\eta \in$ \{0, 0.1, 0.2\}, $\alpha \in$ \{0, 0.1, 0.2\}, $\beta \in$ \{0, 0.1\}, $\gamma \in$ \{0.1, 0.2\}, $m \in$ \{6, 9, 12\}.
	\item
	VGCN: $\bar{\rho}_1 \in$ \{0.25, 0.5, 0.75, 0.99\}, $\bar{\rho}_0=10^{-5}$, $\tau_0 \in $ \{0.1, 0.5\}, $\tau \in$ \{0.1, 0.5\}, $\beta \in$ \{$10^{-4}$, $10^{-3}$, $10^{-2}$, 1\}. Sampling time is 3. Maximum number of training epochs is 5000.
\end{itemize}

\begin{figure}[tb!]
\centering
\subfigure[$\lambda$]{
\begin{minipage}[t]{0.45\linewidth}
\centering
\includegraphics[width=0.99\textwidth,angle=0]{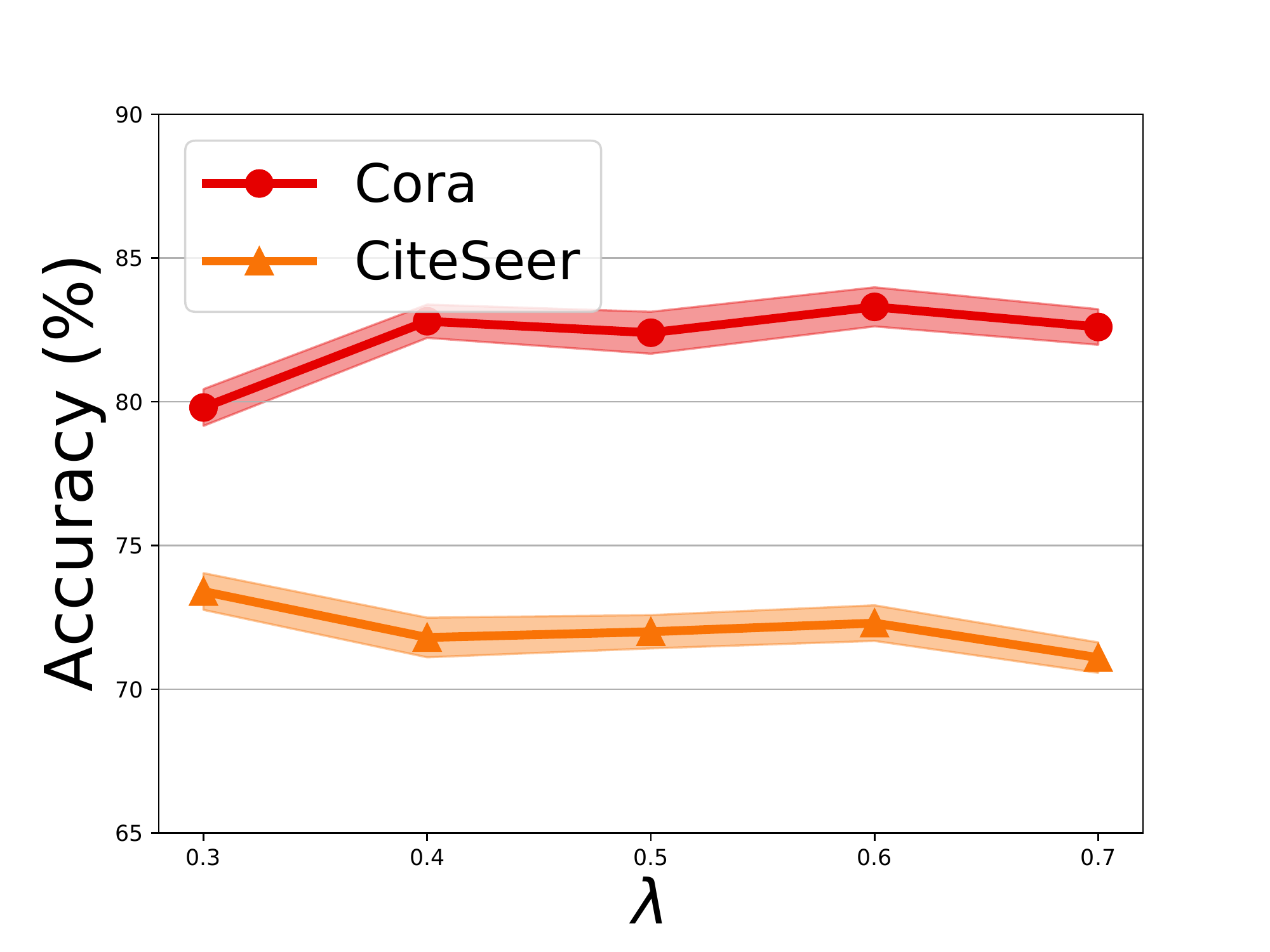}
\label{fig-exp-abl-d}
\end{minipage}%
}%
\subfigure[$P$]{
\begin{minipage}[t]{0.45\linewidth}
\centering
\includegraphics[width=0.99\textwidth,angle=0]{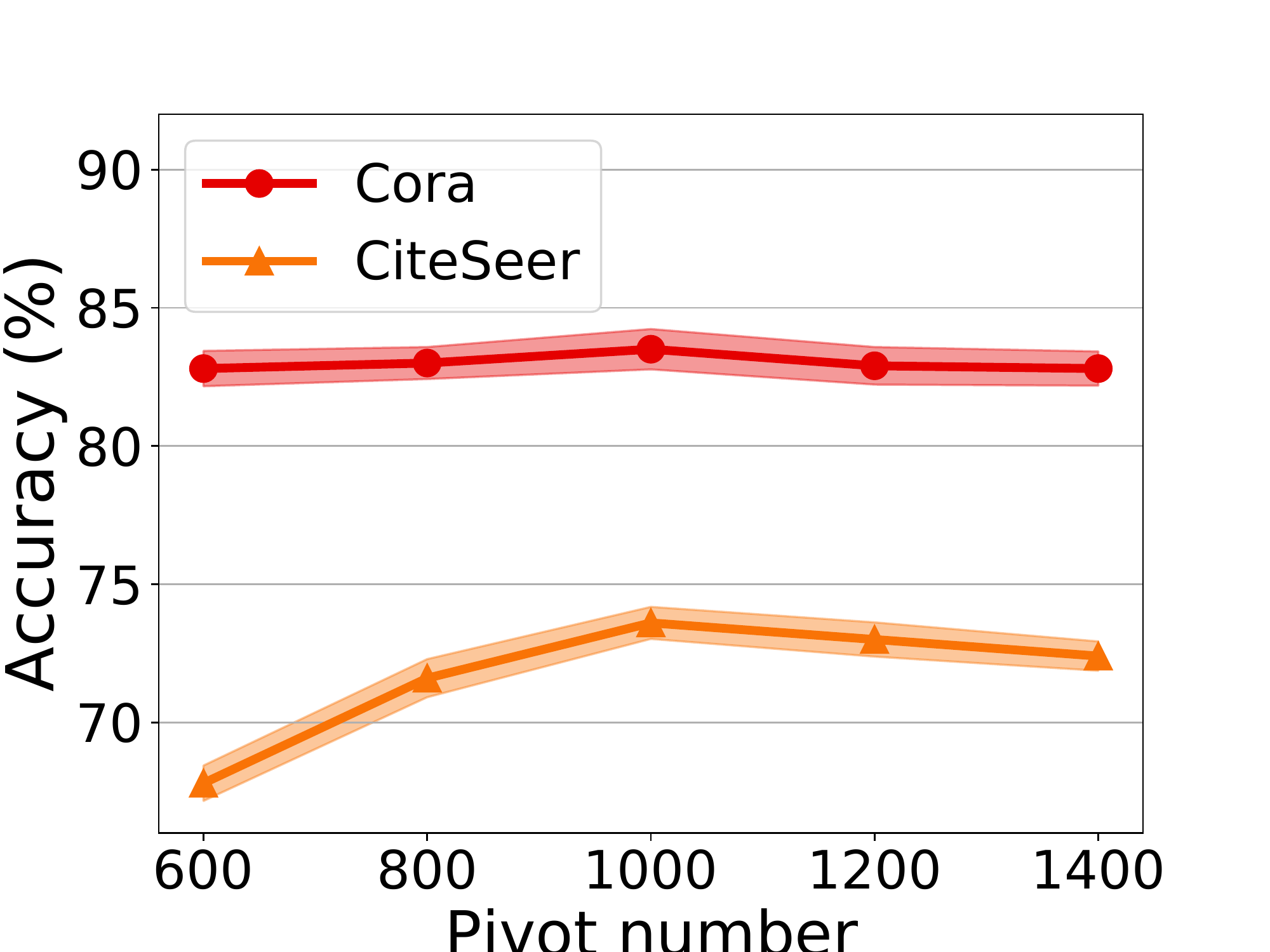}
\label{fig-exp-abl-D}
\end{minipage}%
}%
\caption{Hyper-parameter sensitivity analysis on the weight of input graphs $\lambda$ and pivot number $P$.}
\label{fig-para}
\end{figure}

\begin{figure}[t]
	\centering
	\subfigure[CiteSeer]{
		\includegraphics[width=0.23\textwidth]{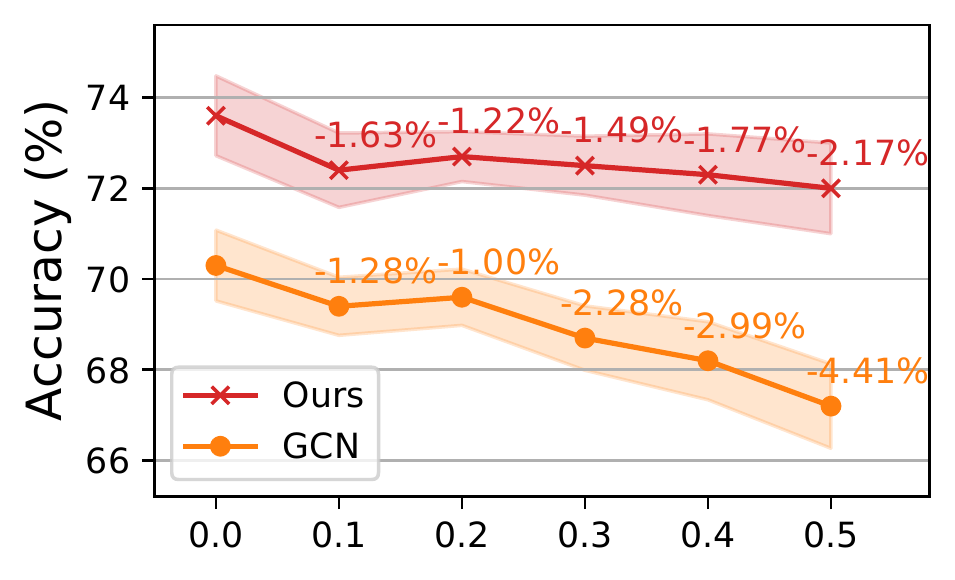}}
	\subfigure[Johns Hopkins55]{
		\includegraphics[width=0.23\textwidth]{figs/e_jh.pdf}}
	\caption{Performance comparison of \mymodel and GCN w.r.t. randomly removing certain ratios of edges in input graphs.}
	\label{fig:edge}
\end{figure}

\section{More Experimental Results}\label{appx-moreres}
\begin{table}
	\begin{center}
	\caption{Comparison between one-layer GCN and MLP as transformation network before structure learning module.}
	\label{tab:linear}
	\begin{tabular}{lcc}
		\toprule
		Dataset & \mymodel-GCN	& \mymodel-MLP \\
		\midrule
		Cora & 83.2 ± 0.4 & 82.5  ± 0.5 \\
		CiteSeer & 73.8 ± 0.9 & 73.3 ± 0.7\\
		PubMed & 79.6 ± 0.7 & 79.6 ± 0.7\\
		Amherst41 & 68.1 ± 1.3 & 68.3 ± 1.5 \\
		Johns Hopkins55 & 71.8 ± 0.7 & 72.1 ± 0.9\\
		Cornell5 & 70.5 ± 1.0 & 69.5 ± 1.0\\
		Reed98 & 67.3 ± 1.2 & 65.9 ± 1.4\\
		\toprule
	\end{tabular}		
	\end{center}	
\end{table}

We compare with using MLP and GCN, respectively, as the transformation network before structure learning module and report the results in Table~\ref{tab:linear}.
In summary, these two methods are of equal competence, which suggests that \mymodel is not sensitive to the transformation network used for converting node features with various dimensions into embeddings with a shared dimension. This also implies that simple neural architectures, e.g. MLP and GCN, could provide enough capacity for extracting the information in input observation, which is leveraged by the shared graph structure learner to discover generalizable patterns in optimal message-passing structure.

We also provide more results of edge deletion experiments in Fig. \ref{fig:edge}. We randomly remove 10-50\% edges of target graphs respectively, and then apply \mymodel and GCN. The results demonstrate that \mymodel is more immune to edge deletion. This is due to our structure learner's ability for learning new structures, making it less reliant on initial graph structures and more robust to attack on input edges.

\end{document}